\documentclass[journal,twoside,web]{ieeecolor}
\usepackage{generic}
\usepackage{cite}
\usepackage{amsmath,amssymb,amsfonts}
\usepackage{algorithmic}
\usepackage{graphicx}
\usepackage{textcomp}

\usepackage{gensymb}
\usepackage{multirow}
\newcommand*\sref[1]{\ref{#1}} 

\usepackage{hyperref}

\iffalse

\usepackage[normalem]{ulem} 
\newcommand\reduline{\bgroup\markoverwith{\textcolor{red}{\rule[0.5ex]{2pt}{2pt}}}\ULon}

\newcommand{\rgerase}[1]{\reduline{#1}}
\newcommand{\rgerasenoline}[1]{#1}

\else

\newcommand{\rgerase}[1]{}
\newcommand{\rgerasenoline}[1]{}

\fi

\def\BibTeX{{\rm B\kern-.05em{\sc i\kern-.025em b}\kern-.08em
    T\kern-.1667em\lower.7ex\hbox{E}\kern-.125emX}}
\begin{document}
\title{Pose Estimation of Periacetabular Osteotomy Fragments with Intraoperative X-Ray Navigation}
\author{Robert B. Grupp,
			Rachel A. Hegeman,
			Ryan J. Murphy,
			Clayton P. Alexander,
			Yoshito Otake,\\
			Benjamin A. McArthur,
			Mehran Armand,
			and Russell H. Taylor, \IEEEmembership{Life Fellow, IEEE}
\thanks{This work was supported by NIH/NIBIB grant R21EB020113, MEXT/JSPS KAKENHI 26108004, and Johns Hopkins University Internal Funds.}
\thanks{R. B. Grupp and R. H. Taylor are with the Department of Computer Science, Johns Hopkins University, Baltimore, MD, USA. (e-mail: grupp@jhu.edu, rht@jhu.edu).}
\thanks{R. A. Hegeman is with the Research and Exploratory Development Department, Johns Hopkins University Applied Physics Laboratory, Laurel, MD, USA. (e-mail: Rachel.Hegeman@jhuapl.edu) }
\thanks{R. J. Murphy was with the Research and Exploratory Development Department, Johns Hopkins University Applied Physics Laboratory, Laurel, MD, USA.
He is now with Auris Surgical Robotics.}
\thanks{C. P. Alexander is with the Department of Orthopaedic Surgery, Johns Hopkins Medicine, Baltimore, MD, USA. (e-mail: calexander@jhmi.edu)}
\thanks{Y. Otake is with the Graduate school of Information Science, Nara Institute of Science and Technology (NAIST), Nara, Japan (e-mail: otake@is.naist.jp).}
\thanks{B. A. McArthur was with Washington Orthopedics and Sports Medicine, Chevy Chase, MD, USA.
He is now with the Dell Medical School, University of Texas, Austin, TX, USA
and
Texas Orthopedics, Austin, TX, USA.}
\thanks{M. Armand is with the Research and Exploratory Development Department, Johns Hopkins University Applied Physics Laboratory, Laurel, MD, USA
and the Department of Mechanical Engineering, Johns Hopkins University, Baltimore, MD, USA. (e-mail: Mehran.Armand@jhuapl.edu)}
\thanks{This paper has supplementary downloadable material available at http://ieeexplore.ieee.org, provided by the authors (File size: 111 MB).}
\thanks{Copyright (c) 2019 IEEE. Personal use of this material is permitted. However, permission to use this material for any other purposes must be obtained from the IEEE by sending an email to pubs-permissions@ieee.org. DOI: \href{https://doi.org/10.1109/TBME.2019.2915165}{10.1109/TBME.2019.2915165}}}
\maketitle
\begin{abstract}
\textit{Objective:}
State of the art navigation systems for pelvic osteotomies use optical systems with external fiducials.
We propose the use of X-Ray navigation for pose estimation of periacetabular fragments without fiducials.
\textit{Methods:}
A 2D/3D registration pipeline was developed to recover fragment pose.
This pipeline was tested through an extensive simulation study 
and 6 cadaveric surgeries.
Using osteotomy boundaries in the fluoroscopic images, the preoperative plan is refined to more accurately match the intraoperative shape.
\textit{Results:}
In simulation, average fragment pose errors were 1.3\degree/1.7 mm when the planned fragment matched the intraoperative fragment,
2.2\degree/2.1 mm when the plan was not updated to match the true shape,
and 1.9\degree/2.0 mm when the fragment shape was intraoperatively estimated.
In cadaver experiments, the average pose errors were 2.2\degree/2.2 mm, 3.8\degree/2.5 mm, and 3.5\degree/2.2 mm when registering with the actual fragment shape, a preoperative plan, and an intraoperatively refined plan, respectively.
Average errors of the lateral center edge angle were less than 2\degree\ for all fragment shapes in simulation and cadaver experiments.
\textit{Conclusion:} The proposed pipeline is capable of accurately reporting femoral head coverage within a range clinically identified for long-term joint survivability.
\textit{Significance:} Human interpretation of fragment pose is challenging and usually restricted to rotation about a single anatomical axis.
The proposed pipeline provides an intraoperative estimate of rigid pose with respect to all anatomical axes, is compatible with minimally invasive incisions, and has no dependence on external fiducials. 
\end{abstract}
\section{Introduction}
\label{sec:introduction}
%
\IEEEPARstart{D}{evelopmental} hip dysplasia (DDH) is a congenital condition which may cause greater contact pressures between the femoral head and acetabular cartilage due to a deformed acetabulum or femoral head.
These conditions may cause pain and limit the amount of physical activity an individual may perform.
When DDH is not addressed, severe arthritis by age 30 is typical, and requires surgical intervention \cite{murphy1995prognosis}. 

\begin{figure}[!t]
\centerline{\includegraphics[width=\columnwidth]{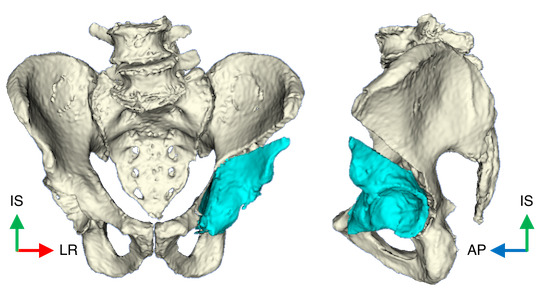}}
\caption{Example of periacetabular osteotomies and a fragment reposition.
This is taken from the cadaver experiments and represents the ground truth fragment pose.
The manually segmented fragment shape is shown.
See Fig. \ref{fig:frag_xrays} for the corresponding intraoperative fluoroscopic images.
The orientations of each anatomical axis for the anterior pelvic plane are also depicted: left/right x axis (LR), inferior/superior y axis (IS), and anterior/posterior z axis (AP).}
\label{fig:frag_view_3d}
\end{figure}

%
In 1984, the periacetabular osteotomy (PAO) was developed by Ganz for the purpose of treating DDH\cite{ganz1988new}.
Standing AP and false profile radiographs are used to preoperatively assess the condition and develop an initial surgical plan.
Osteotomies along the pubis, ilium, ischium, and posterior column of the pelvis are performed, typically with fluoroscopic guidance \cite{ganz1988new}.
The ischial and posterior cuts present significant challenges, including using difficult to interpret fluoroscopic views, potentially affecting the Sciatic nerve, and risking joint breakage due to the osteotome's proximity to the acetabulum.
After completing the osteotomies, the acetabular fragment is freed from the pelvis and repositioned to increase femoral coverage and improve joint contact pressure \cite{ganz1988new,armand2005outcome,armiger2009three,niknafs2013biomechanical,hipp1999planning} (Fig. \ref{fig:frag_view_3d}).

Mentally resolving the 3D pose of the acetabular fragment using intraoperative fluroscopy is challenging, especially for novice surgeons \cite{troelsen2009surgical}.
An acceptable repositioning is often determined using the lateral center edge (LCE) angle \cite{wiberg1939studies}, which estimates the amount of lateral femoral head coverage provided by the acetabulum.
In order to achieve long-term survivability of the joint, a LCE angle between $30\degree-40\degree$ is desired \cite{hartig2012factors}.
Mentally resolving the LCE angle from 2D radiographs, without any other tools, was shown to have a large variance \cite{troelsen2010assessment}.

%
To address the challenges with performing PAO, several computer assisted systems using optical tracking technology have been developed; e.g. \cite{langlotz1998computer,murphy2015development,mayman2002kingston,liu2014computer,akiyama2010computed,takao2017comparison}.
However, intraoperative optical tracking systems are not yet standard equipment in most operating rooms, and have several technical disadvantages, such as a sensitivity to occlusion and a limited workspace.
Moreover, the need to digitize points on the ilium and iliac crest for registration of the patient's pelvis to a preoperative model, requires more invasive incisions than typically needed for a PAO \cite{murphy2015development,troelsen2008new,troelsen2008comparison}.
These reasons, and the universal availability of intraoperative fluoroscopic imaging, provide motivation for the use of an X-Ray based navigation system.
In place of specialized 3D tracking devices, computer-assisted X-Ray navigation systems use the fluoroscopic or radiographic imagers already present during many interventions \cite{belei2007fluoroscopic,chen2013automatic,dennis2005vivo,gong2011multiple,gueziec1998anatomy,joskowicz1998fracas,mahfouz2003robust,markelj2012review,otake2012intraoperative,otake2016robust,penney1998comparison,sheng2015automatic,yao2000ac,yi2018robotic}.
Since the acetabular fragment is created intraoperatively, an accurate model of the fragment shape is not available, and invalidates the assumptions used in existing multiple-object registration methods \cite{gong2011multiple,joskowicz1998fracas,otake2016robust}.

\begin{figure}[!t]
\centerline{\includegraphics[height=\columnwidth]{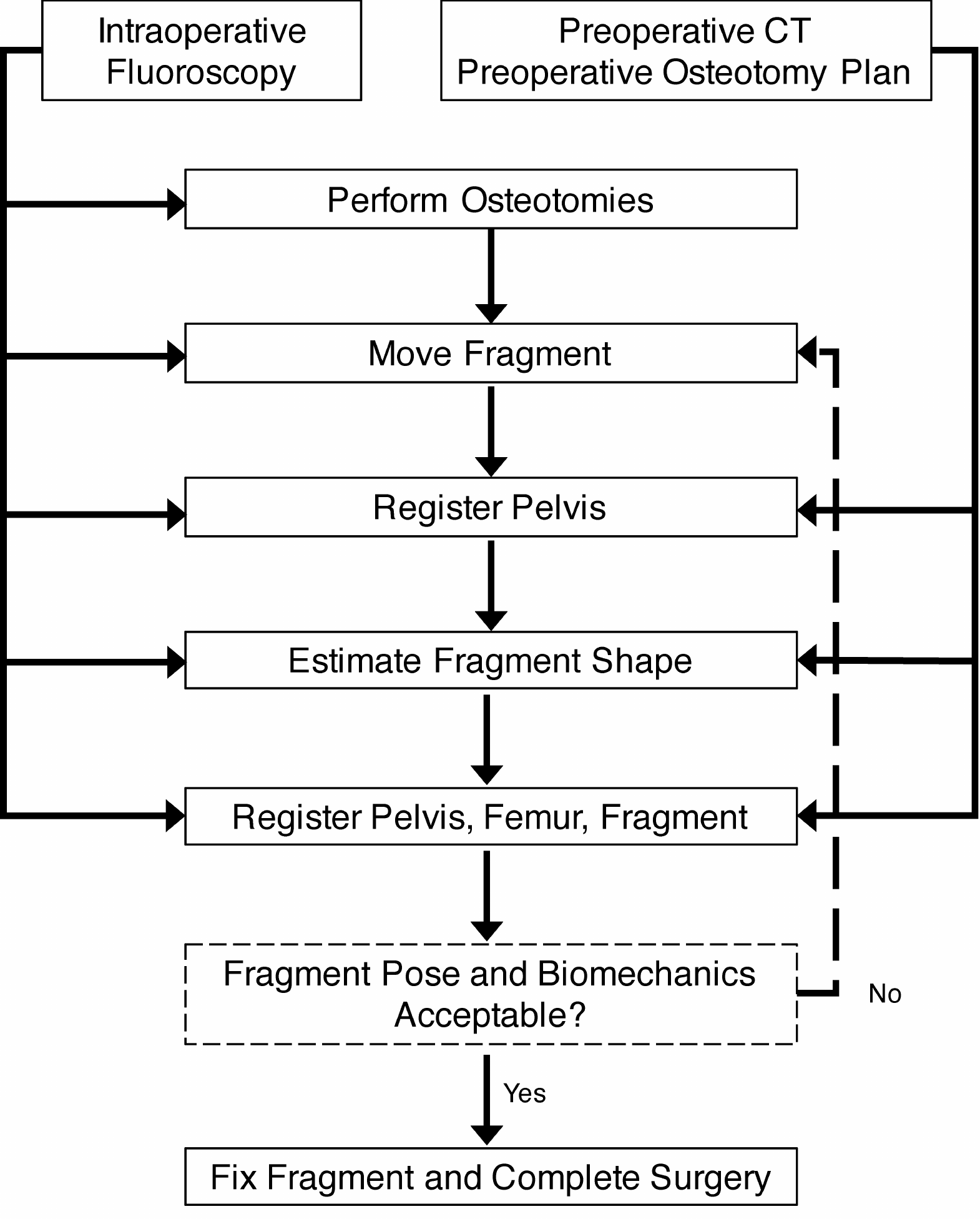}}
\caption{High level workflow detailing the steps performed during the surgery and the data required for each. The registration and shape estimation steps are the primary focus of this paper.}
\label{fig:workflow}
\end{figure}

We propose a fiducialess approach that performs multiple-object, multiple-view, 2D/3D X-Ray to CT registration to resolve the pose of an acetabular fragment with respect to the anterior pelvic plane (APP).
After the clinician performs a repositioning of the fragment, the pelvis and fragment are registered using several fluoroscopic images, and the fragment's pose and LCE angle are reported.
The fragment shape is estimated after registering the patient's non-acetabular portion of the pelvis.
Cut lines present in the 2D images are used to approximate the ilium and pubis osteotomies, while the ischial and posterior osteotomies remain set according to a preoperative plan.
The pose and LCE values may be interpreted by the clinician to determine whether the fragment is in an acceptable pose, or whether further adjustment is needed.
When the fragment needs to be moved, the current pose estimate may help determine the direction in which the fragment should next be adjusted.
Additionally, but beyond the scope of this work, biomechanical indicators may also be computed once the fragment's relative pose is known \cite{armand2005outcome,armiger2009three,niknafs2013biomechanical,hipp1999planning}.
The high-level workflow of our proposal is shown in Fig. \ref{fig:workflow}.

We believe this is the first computer assisted system for intraoperatively tracking an acetabular fragment with X-Ray navigation without the use of artificial fiducial objects.
Compared to other multiple-view registration approaches in literature, our approach does not require a tracked, encoded, or motorized, C-Arm.
Unlike existing multiple-object registration solutions, an inaccurate model of the fractured shape is allowed and a prior distribution over fragment shapes is not required.
Moreover, non-standard equipment is not required to perform the surgery and no burdensome steps are added to the operating workflow.

We have evaluated the proposed method with a large simulation study and six cadaver surgeries.
In both experiments we report fragment pose and LCE errors when knowing the true fragment shape, a preoperatively planned shape, and an intraoperatively estimated shape.
To evaluate the performance of our method, we examine rotation and translation components of the pose differences from ground truth, as well as the error in LCE angle measurement.
Since it is expected that fragment pose and LCE angle errors will increase as the shapes used for registration match intraoperative shapes less, analysis of the simulation and cadaver experiments will determine if the proposed methods are sufficiently accurate to assist with intraoperative evaluation of dysplasia.
%
\section{Related Work}\label{sec:rel_work}
Initial work on computer aided systems for pelvic osteotomy focused on performing the actual osteotomies, rather than tracking the mobilized acetabular fragment. 
Langlotz demonstrated the use of preoperative CT and optically tracked surgical instruments to assist with PAOs, but did not provide estimates on the relocated pose of the acetabular fragment \cite{langlotz1998computer,langlotz1997first}.
Mayman used optical tracking to intraoperatively place guiding screws on the patient's pelvis and matching osteotomies from a preoperative plan, but did not provide intraoperative tracking of the fragment \cite{mayman2002kingston}.
Akiyama also tracked osteotomes during a curved periacetabular osteotomy (CPO) \cite{naito2005curved} using optical tracking \cite{akiyama2010computed}.
In \cite{radermacher1998computer}, Radermacher constructed patient specific cutting guides for use in a triple osteotomy \cite{tonnis1994triple}, however the pose of the fragment was not reported during the procedure.
Similarly, Otsuki developed patient specific cutting guides to perform CPO and did not track the pose of the acetabular fragment \cite{otsuki2013developing}.

Also using preoperative CT and optical tracking devices, Murphy did not track the ostoetomes, but instead tracked the acetabular movements and computed intraoperative biomechanics \cite{murphy2016clinical,murphy2015development}.
Through repeated digitizations of points on the acetabular fragment body, the fragment was tracked and the appropriate biomechanical properties were presented to the clinician \cite{murphy2015development}.
Fragment pose errors of $1.4-1.8 \degree$ and $1.0 - 2.2$ mm were reported.
The process of manually digitizing the bone bur points during each fragment reposition adds a small amount of time to the overall procedure and may be subject to some error \cite{murphy2016clinical}.
Liu also developed a system for preoperative PAO planning and intraoperative tracking of the fragment using optical tracking and a separate rigid body attached to the fragment \cite{liu2014computer,liu2016periacetabular}.
However, fixing a separate rigid body to the fragment is not necessarily practical when using a state-of-the-art, minimally invasive, approach such as \cite{troelsen2008new} and \cite{troelsen2008comparison}.
For rotational acetabular osteotomies (RAO) \cite{ninomiya1984rotational}, Takao used an optically tracked system to monitor the osteotomes and fragment movement \cite{takao2017comparison}.
The fragment pose was intraoperatively estimated by digitizing the anterior edge of the acetabulum, which resulted in some difficulty distinguishing between rotation and translation.

%
X-Ray navigation has been used to assist with the reduction of traumatic bone fractures, by providing real-time 3D visualizations of relocated bone fragments \cite{joskowicz1998fracas} and feedback corresponding to a 3D preoperative plan \cite{gong2011multiple}.
Methods to automatically annotate intraoperative images have been used to avoid wrong-level spinal surgery \cite{sheng2015automatic} and mitigate the malpositioning of surgical implants \cite{belei2007fluoroscopic}.
Using intraoperative X-Ray imaging, a surgical robot may be guided into an optimal location for the milling \cite{gueziec1998anatomy,yao2000ac}, or drilling \cite{yi2018robotic}, of bone.
Automatic 3D visualization and kinematic analysis of the wrist \cite{chen2013automatic} and knee \cite{mahfouz2003robust,dennis2005vivo} have also been demonstrated with X-Ray navigation.
To our knowledge, no existing method based on X-Ray navigation, without fiducials, is able to localize a bone fragment without accurate preoperative knowledge or a statistical prior of the fragment shape.

At the core of an X-Ray based navigation system is a 2D/3D registration algorithm \cite{markelj2012review}.
The goal of 2D/3D registration is to determine the pose of 3D objects with respect to a 3D coordinate frame using a series of 2D X-Ray images.
Typically, a preoperative 3D model, such as a CT scan is used to represent the patient, and the information contained in the X-Ray image is used to determine the pose of
the patient with respect to the intraoperative X-Ray imager.
The majority of 2D/3D X-Ray registration methods may be classified as either ``intensity-based'' or ``feature-based,'' however we limit discussion to intensity-based methods in this paper.

%
%
Intensity-based registration performs an optimization over the relevant pose parameters, using an objective function that compares simulated radiographs, commonly referred to as digitally reconstructed radiographs (DRRs), with the intraoperative image \cite{penney1998comparison}.
The comparison is performed using a mathematical construct known as a similarity measure \cite{penney1998comparison}.
Due to the differences in X-Ray energy between preoperative CT and intraoperative fluoroscopy, the most effective similarity metrics compare the 2D gradients of a measured radiograph and a DRR, such as normalized cross-correlation between the Sobel gradient images (Grad-NCC) \cite{penney1998comparison}.
Robustness to metallic objects and bone fractures may be improved by taking a weighted sum of similarity measures computed in local regions of interest \cite{knaan2003effective,grupp2018patch}.
Registration with multiple 2D views, and known relative poses between each view, is accomplished by creating DRRs at each view and summing the similarity scores for each view \cite{otake2012intraoperative}.
In order to register multiple objects with known shape, each object may be treated as a separate volume, and DRRs for each object are summed together to create a single DRR \cite{otake2016robust}.
The registration problem for $N$ object poses: $\theta_1, \dots, \theta_N$, with $M$ intraoperative views: $I_1 \dots I_M$, a pre-operative CT: $V$, a DRR operator: $\mathcal{P}$, similarity metric: $\mathcal{S}$, and regularizer over plausible poses: $\mathcal{R}$, is concisely stated in \eqref{eq:intensity_regi_opt_prob}.
\begin{equation} \label{eq:intensity_regi_opt_prob}
  \min_{\theta_1, \dots, \theta_N \in SE(3)} \sum_{m = 1}^M \mathcal{S}\left(I_m, \sum_{n=1}^N \mathcal{P} \left( V; \theta_n \right) \right) + \mathcal{R} \left( \theta_1, \dots \theta_N \right)
\end{equation}

%
With the advent of general purpose GPU programming resources, Otake was able to efficiently form many DRRs simultaneously and use a state of the art ``Covariance Matrix Adaptation: Evolutionary Search'' (CMA-ES) optimization strategy \cite{hansen2001completely} to carry out registration of a single femur using three views in under 22 seconds \cite{otake2012intraoperative}.
Relative pose information was computed using an external fiducial for a non-motorized C-Arm and was preoperatively calibrated when using a motorized C-Arm.

Several groups have demonstrated registration of multiple objects with intensity-based objective functions and accurate shape models, or with a statistical prior of the shape distributions.

In \cite{otake2016robust}, Otake's framework was extended to multiple objects for the knee joint (distal femur, patella, proximal tibia) tracking with bi-plane fluoroscopy.
Initial registration times at the start of each sequence for the femur, tibia, and patella bones were between 2 and 5 minutes.
All femur and tibia poses were estimated within $2 \degree$ and $2$ mm, and 74\% of patella poses were estimated within the same thresholds.


Gong proposed to use intensity-based registration to intraoperatively estimate the position of bone fragments resulting from a distal radius fracture \cite{gong2011multiple}.
The approach requires preoperative knowledge of the bone fragment shapes and uses a preoperative, but post-trauma, CT scan \cite{gong2011multiple}.
Two 3D printed phantoms with synthetic fractures were used to test the method with four fluoroscopic views from a tracked C-Arm.
Target registration errors (TREs) smaller than 3 mm were achieved when using a manual, interactive, initialization of the registration.
Execution times of 3-9 minutes were reported using modest hardware.

In order to localize and determine the shape of carpal bones in the hand, Chen, et al. use a 2D/3D registration of a single fluoroscopic view to 3D statistical shape and pose models of the carpal bones, radius, and ulna \cite{chen2013automatic}.
TREs of $2.45$ mm were reported in simulation, and TREs from $0.93 - 2.37$ were reported in the flouroscopic experiments.
Registration times were approximately 3 minutes per frame.

Although related by the motivation to track multiple objects with intensity-based registration, the aforementioned works do not provide a complete solution for the localization of an acetabular fragment.
Most importantly, the fragment shape is not completely known pre-operatively and, to our knowledge, no statistical priors of pelvic osteotomies exist.
In contrast to the fluoroscopic data collected of the phantom in \cite{gong2011multiple}, the fluoroscopic images of a human have more heterogeneous hard and soft tissue distributions, usually resulting in a more challenging registration.
Additionally, objects such as metallic screws, wires, or tools may confound a registration strategy, and were not present in the fluoroscopic views used by \cite{otake2016robust,gong2011multiple,chen2013automatic}.

Without any external fiducial objects or optical tracking systems, the methods proposed in this paper intraoperatively estimate an acetabular fragment's pose using X-Ray navigation.
The pose estimation problem is solved by extending previous 2D/3D X-Ray registration techniques to partially estimate the acetabular fragment shape during surgery, and then localize the pelvis and fragment without a tracked, motorized, or encoded, X-Ray imager.
Furthermore, our method aims to provide intraoperative feedback in a reasonable timeframe, with computation times on the order of seconds.
%
\section{Materials and Methods}
\label{sec:methods}
\subsection{Preoperative Processing and Planning}\label{sec:preop_proc_plan}
Lower torso preoperative CT images were acquired and resampled to have 1 mm isotropic voxel spacing.
Segmentation of the pelvis, and left and right femurs was performed using an automatic method \cite{krvcah2011fully}, followed by manual touch up.
The manual touch up was occasionally necessary to distinguish between the acetabulum of a pelvis and the corresponding femoral head.
Using manually annotated landmarks, the APP was computed as described by \cite{nikou2000description} and the origin was relocated to the ipsilateral femoral head.
The APP coordinate axes are aligned with the left/right (LR), inferior/superior (IS), and anterior/posterior (AP) anatomical axes.
An example of the APP axis orientations is shown in Fig. \ref{fig:frag_view_3d}.
In order to report LCE angles intraoperatively, the most lateral points of the acetabulum are digitized from coronal slices of the preoperative CT volume.

Other landmarks identified for registration purposes were the medial and inferior points of maximal curvature on the obturator foramen, the greater sciatic notch, and inferior symphysis.
To allow for accurate initialization of full pelvis registration in the presence of a mobilized fragment, the registration landmarks should not be located on portions of the pelvis that may become fragment.

A preoperative plan of the osteotomies is created by manual selection of several landmarks on the pelvis surface and cutting planes are fit to pass through the landmarks.
Next, a label map is computed that indicates whether a voxel represents the acetabular bone fragment, the non-fragment pelvis, a portion of bone removed due to the chiseling action, left femur, or right femur.
The osteotomy action is simulated by moving a virtual chisel, of width 1 mm, along the virtual cutting planes defined by the preoperative plan.
Starting from the original segmentation, all pelvis labels contained within the convex hull defined by the virtual cutting planes are marked as a candidate fragment voxel.
Furthermore, if a candidate fragment voxel is within a chisel width of any virtual cut plane, it is also marked as a candidate ``cut'' voxel.
The set of candidate fragment voxels is divided into connected components, with the largest component kept as candidate fragment voxels, and the others reverted to pelvis labels.
Of the remaining candidate fragment voxels, any marked as candidate cut voxels are labeled as cuts, and are otherwise labeled as fragment.
Tight bounding boxes are computed about the pelvis labels, fragment labels, and femur labels to create sub-volumes for each object. 
%
\subsection{Intraoperative Registration Strategy}\label{sec:gen_regi}
\begin{figure}[!t]
\centerline{\includegraphics[width=\columnwidth]{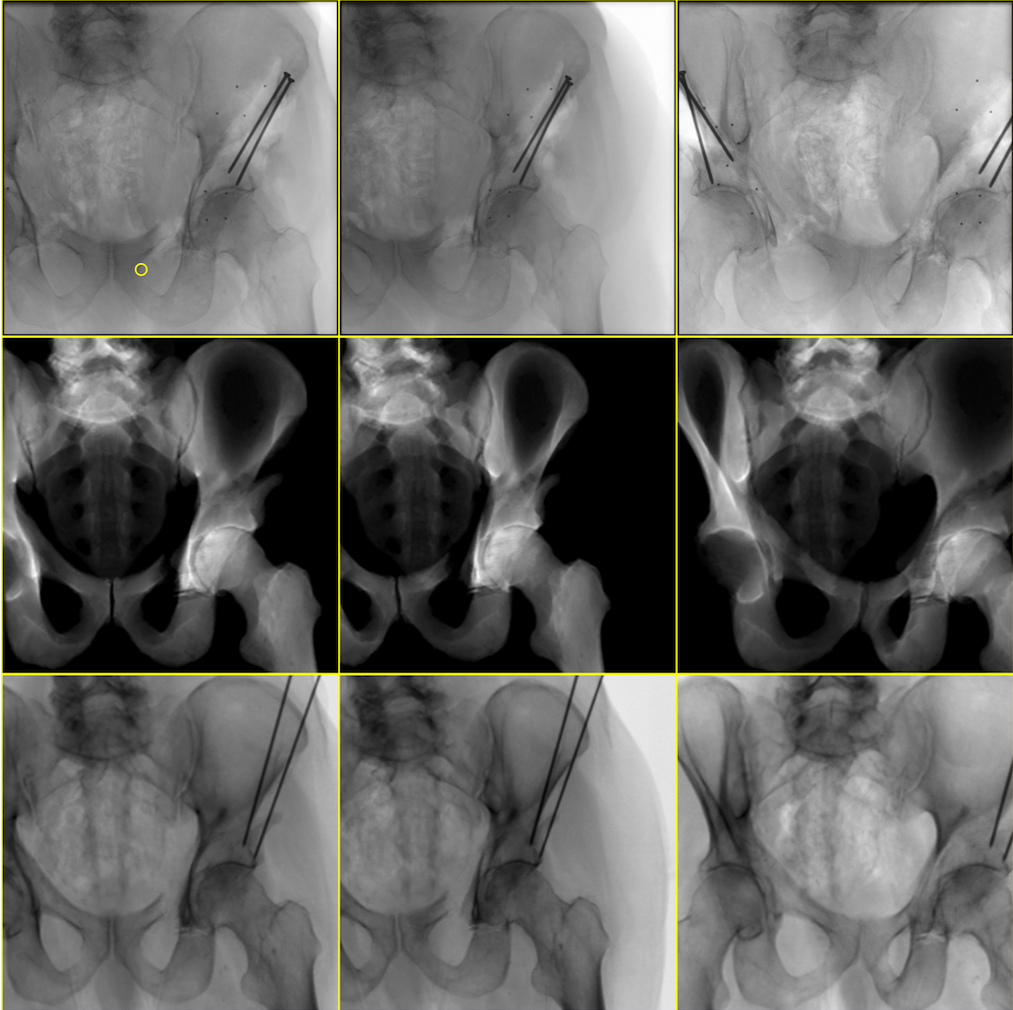}}
\caption{Top Row: Intraoperative fluoroscopic images used for registering the fragment on the left side of a cadaveric specimen.
Note the four BBs on the mobilized fragment and four BBs on the ilium used for ground truth computation.
The left image shows an approximate AP view and the yellow circle indicates the single-landmark used for initialization of the intraoperative pipeline.
Middle Row: DRRs created as part of the intensity-based registration using manual segmentation of the fragment; the final pose estimates are shown.
Prior to performing any similarity comparisons, the images in the top-row will be log-corrected.
Bottom Row: An example of simulated fluoroscopic data generated using the protocol of the simulation study.
The image includes soft-tissue and inserted K-wires, with object poses determined by the registrations from the middle row.
Despite the visual similarity with the top row, the challenge of visually determining a fragment's pose is confirmed by an approximate 3\textdegree/3 mm pose error difference from the top row's ground truth.
See Fig. \ref{fig:frag_view_3d} for the ground truth 3D view.}
\label{fig:frag_xrays}
\end{figure}
Our registration strategy is based on the methods described in \cite{otake2012intraoperative} and \cite{otake2016robust}.
However, the methods proposed here do not rely on external information to recover the multiple-view geometry,
provide a quick single-landmark initialization,
use a similarity metric robust to mismatches between the preoperative models and intraoperative reality,
perform a faster local search at higher resolution levels,
and attempt to recover the shape of intraoperatively created bone fragments without an additional CT.

DRRs are formed by ray casting through each object's attenuation volume with a ray step size of 1 mm and trilinear interpolation.
Grad-NCC scores are computed throughout image patches, with the mean value used as the similarity measure for all intensity-based registrations in this work \cite{grupp2018patch}.
Ray casting and similarity metric computation are computed on a GPU through the use of OpenCL \cite{stone2010opencl}, while optimization logic is conducted on the CPU.
Image parameters and the projection geometry are derived from a Siemens CIOS Fusion C-Arm with 30 cm flat panel detector.
Using DICOM metadata populated by a CIOS Fusion, we created a na{\"i}ve set of intrinsic parameters for the C-Arm: image dimensions of $1536 \times 1536$, isotropic pixel spacing of $0.194$ mm/pixel, a source-to-detector distance of 1020 mm, no distortion, and a principal point at the center of the image.

We use two 2D image resolution levels in this work, starting from coarse ($8\times$ downsampling in each 2D dimension) and moving to a finer resolution ($4\times$ downsampling).
At the coarsest resolution level, all optimizations are conducted with the CMA-ES strategy, and the Bounded Optimization by Quadratic Approximation (BOBYQA) strategy \cite{powell2009bobyqa} is used at the second level.
The CMA-ES search strategy is a derivative free method which requires a large number of objective function evaluations per iteration, but approximates a traditional second-order algorithm while maintaining robustness to local minima \cite{hansen2001completely}.
Since the BOBYQA strategy requires only a single objective function evaluation per iteration, it has significantly faster runtimes compared to CMA-ES for this work.
Even though BOBYQA does not provide the same robustness to local minima as CMA-ES, we justify its use at the second resolution level due to its efficiency, and because we assume that pose estimates after the first resolution level lie within a smooth, convex, region about the true minima.

The C-Arm is not required to be motorized, encoded, or capable of reporting relative pose information.
In order to obtain the relative pose information required by the multi-object fragment registration, we treat the patient's anatomy as a ``fiducial'' object.
Single-view registrations of the pelvis are performed for each view and the pelvis coordinate frame is then used as the C-Arm world frame.

Assuming the first fluoroscopic view is approximately AP, a single landmark is manually annotated in 2D to initialize pelvis registration.
The 3D landmark location, with respect to the C-Arm, is estimated by traversing the source-to-detector ray of this landmark $85\%$ of the source-to-detector distance.
Orientation of the pelvis is computed using the APP coordinate frame described in section \ref{sec:preop_proc_plan}.
Translation is recovered by aligning the landmark points in the two coordinate frames.
The point of maximal curvature on the medial portion of the ipsilateral obturator foramen was chosen as the initialization landmark used in this work, and an example annotation is shown in the top left of Fig. \ref{fig:frag_xrays}.
If the first view is not approximately AP, an initial pose estimate of the pelvis may be obtained by annotating more landmarks in 2D and solving the PnP problem \cite{hartley2003multiple}.
Once the initial pose is estimated, a single-view intensity-based registration of the pelvis is performed.

By restricting the possible movements of the C-Arm for subsequent views, initial pelvis pose estimates for the remaining views are automatically obtained through an exhaustive search starting from the registered pose of the pelvis in the first view.
More precisely, we limit the geometries of the remaining views to only differ from the first view by a rotation along the C-Arm's orbit.
DRRs are computed by adjusting the registered pelvis pose of the initial view by C-Arm orbital rotations of $\pm 90 \degree$ in $1 \degree$ increments, for a total of 181 DRRs.
Similarity computations are made between the second and third fluoroscopic views, and the poses corresponding to the best similarity scores are used to initialize intensity-based registrations.
An offline calibration process was conducted to determine center of orbital rotation and rotation axis for the CIOS Fusion.

Prior to any further registrations, the preoperative plan of the fragment may be refined as described in section \ref{sec:method_frag_shape}.

Once the single view pelvis registrations are complete, the multiple-view geometry is recovered using the pelvis coordinate frame and the multiple-object registration is conducted.
Specifically, we attempt to solve problem \eqref{eq:intensity_regi_opt_prob} through successive optimizations of individual objects, followed by a simultaneous optimization.
At each resolution level, the pelvis pose estimate is refined starting from its current estimate.
The pose of the pelvis is kept fixed, while an optimization over the pose of the femur is performed.
Keeping the poses of the pelvis and femur fixed, the pose of the acetabular fragment is estimated.
At the coarse level, the femur and fragment poses are initialized with the current estimate of the pelvis pose, while previous poses are used for initialization at the second level.
At the second level, a ``simultaneous'' optimization is performed after the ``sequential'' optimizations; which optimizes over all object poses simultaneously.
An example of three views used for a fragment localization is shown in Fig \ref{fig:frag_xrays}.

The registration process produces pose estimates of the extrinsic C-Arm frame ($C$) with respect to the pelvis ($PV$) and fragment ($FV$) volumes: $T_C^{PV}$ and $T_C^{FV}$.
The relative pose of the fragment with respect to the APP is computed as:
\begin{equation} \label{eq:rel_pose}
	\Delta_{{APP}} = T_{PV}^{{APP}} T_C^{FV} T_{PV}^C T_{{APP}}^{PV}.
\end{equation}
$T_{APP}^{PV}$ is obtained when estimating the APP from preoperative landmarks and maps points in the APP to the preoperative pelvis and fragment volumes.
Using $\Delta_{{APP}}$, the intraoperative fragment movement may be visualized in 3D (see Fig. \ref{fig:frag_view_3d}) and shown to the clinician.
Additional information useful for the assessment of dysplasia may also be displayed, such as the LCE angle and a decomposition of the pose with respect to each anatomical axis.
Similar to the approach in \cite{armiger2007evaluation}, the LCE angle is computed by applying the fragment's relative pose to transform the 3D preoperative lateral trace of the acetabulum.

Given initial estimates for poses of each object, the optimization for each object is performed over the $\mathfrak{se}(3)$ Lie Algebra \cite{murray1994mathematical}, with the $SE(3)$ reference point set from the initialization point.
For a single view registration, an intermediate coordinate frame that is axis aligned with the camera projection frame, with origin at the initial estimate of the femoral head, is used.
For multiple view registrations, the APP with origin at ipsilateral femoral head is used as an intermediate frame during the optimization.
To discourage implausible poses, regularization is applied when using the CMA-ES optimization strategy.
For single view registrations, regularization is applied separately to each translation component and the Euler decompositions of the rotation matrix, whereas the total rotation and translation magnitudes are used for multiple view registrations.
Box constraints are used for all optimizations using the BOBYQA optimization strategy.
Since the pelvis is initially registered with single-views, the optimization parameters used during the multi-view/multi-object registration for the pelvis are set for quicker execution time and tighter constraints in comparison to those used to register the femur and fragment.

Exact parameter values are listed in supplementary section \sref{sec:supp_gen_regi}.
\subsection{Fragment Shape Estimation} \label{sec:method_frag_shape}
Since the fracture line introduced during bone chiseling is dependent on bone quality, the cut does not always follow a planned path.
Therefore, it is not realistic for a clinician to exactly reproduce the planned cuts, even when the osteotome is guided by a navigation system.
To account for the potential uncertainties associated with the osteotomies, we have developed a method to estimate the fragment shape; starting from a preoperative plan and refined from intraoperative fluoroscopic images.

\begin{figure}[!t]
\centerline{\includegraphics[width=\columnwidth]{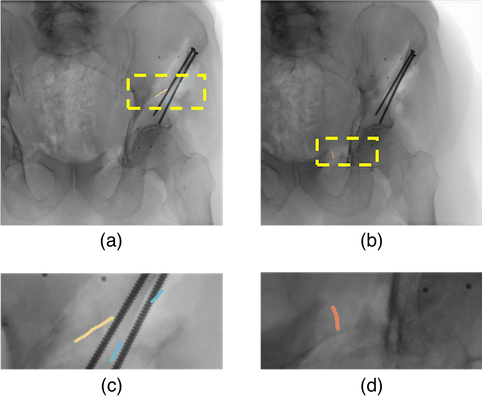}}
\caption{An example of the 2D cut manual annotations from fluoroscopic images corresponding to the case shown in Fig. \ref{fig:frag_view_3d}.
Labels identifying ilium cuts are shown in (a), and labels identifying pubis cuts are shown in (b).
Zoomed displays of the regions of interest in (a) and (b) are shown in (c) and (d), respectively.
In (a) and (c) biege labels indicate rays that enter the pelvis surface when moving from the detector to source, while blue labels indicate rays that exit the pelvis surface.
In (b) and (d), red labels indicate rays that both enter and exit the pelvis surface; the view looks approximately down the cut line.}
\label{fig:cut_2d_annotate}
\end{figure}

For each osteotomy to be estimated, a post-osteotomy fluoroscopic view clearly showing the cut lines is required.
The full pelvis shape is registered to this view and a user manually annotates 2D pixels along the cut lines visible on the main pelvis object.
For each 2D label, a ray is cast from the C-Arm detector towards the X-Ray source, and the 3D intersection points with the pelvis surface are computed.
The osteotomy is estimated by fitting a plane to the recovered 3D points.
When looking approximately down the cut line, the ray is nearly tangent to the entire cut and 3D intersection points entering and leaving the pelvis surface are used.
Otherwise, a label must indicate whether the ray intersects the 3D osteotomy when entering or exiting the pelvis surface;
in this case only the entry or exit intersection point is used.
Fig. \ref{fig:cut_2d_annotate} shows an example of the 2D labeling of ilium and pubis cut pixels.

If an appropriate view and corresponding pelvis registration is not available, then the preoperative plan for that osteotomy is used for shape estimation.
For the ischial and posterior osteotomies, it is challenging to obtain a view that clearly shows the cut lines, and for which a pelvis registration may be successfully performed.
Therefore, only the ilium and pubis osteotomies will be estimated in the following experiments, and the ischium and posterior osteotomies will be set at the planned values.

\subsection{Simulation Study} \label{sec:method_sim_study}
In order to determine the feasibility of this registration strategy, we conducted a large simulation study using the CT scans of 6 non-dysplastic cadaveric specimens.
The group of specimens consisted of 4 male and 2 female subjects, with ages ranging from 57 to 94 years ($81 \pm 14$).
For the left and right sides of each specimen, we created simulated acetabular fragments, simulated fragment movements, and simulated intraoperative fluoroscopic images.

All PAOs were performed bilaterally by a single surgeon (B.A.M.).
In order to create nominal fragment shapes, planes were fit to the osteotomy contours in postoperative CT volumes and then mapped into preoperative CT coordinates.
Random surgeries were simulated by applying random rigid transformations to the cutting planes associated with each nominal osteotomy.
The random adjustments were manually verified to create plausible PAO fragments and any invalid fragments were rejected.
This resulted in 15 simulated fragments for each side of each patient and 180 for the entire study.

Random movements of each fragment and ipsilateral femur were sampled.
Random rotations of the fragment and femur were sampled along with a random translation component for both the fragment and femur.
The translation component is shared, since the fragment and femur tend to move together.
To mimic clinically relevant movements of the fragment and femur, Euler angles and translations about each anatomical axis were sampled.
All transformations are with respect to the APP with appropriate femoral head origin.
Collision detection was performed to ensure that different bones did not overlap in 3D space.
When a collision occurred, the set of transformations was rejected, and another set was sampled.
A total of 20 movements for each fragment were sampled, with a mean rotation of $15.49\degree \pm 6.77\degree$ and mean translation of $4.35 \text{ mm } \pm 1.57 \text{ mm}$.
This yields a total of $3,600$ fragment movements for this study.

Three views are used for each repositioned fragment: a perturbed AP view, followed by two views offset at random rotations about the C-Arm's orbit.
The simulated fluoroscopy incorporates soft-tissues by piece-wise rigidly warping the original CT volume according to the pelvis, femur, and fragment labels as described in \cite{grupp2018patch}.
A temporary fixation of the fragment is simulated by inserting two random K-wires into the volume, each at a random pose intersecting the ilium and fragment.
Using a protocol similar to \cite{markelj2010standardized}, fluoroscopic images are created from this new volume and view geometries.
An example set of simulated fluoroscopic images is shown in the bottom row of Fig. \ref{fig:frag_xrays}.

The registration process proceeds according to the strategy described in \ref{sec:gen_regi}.
Registrations are initialized by simulating a 2D and 3D point picking process for the single landmark strategy.
Using the registration point manually identified in each CT, the corresponding 2D pixel location is computed for each set of fluoroscopic views.
Random noise is added to the 2D and 3D points to simulate variation in human point picking behavior.
The landmarks are offset by a random distance along a random direction.
The offset directions are drawn uniformly from the appropriate (2D or 3D) sphere.
The 3D offset magnitudes are sampled, in mm, from $N(0,3)$, and the 2D magnitudes are sampled, in pixels, from $N(0,7.5)$.
Five different initializations are estimated for each set of fluoroscopic views.

These simulations result in a total of $18,000$ multiple-object/multiple-view registrations that need to be evaluated.
Each registration is run over four scenarios; each with a different fragment shape used during the registration.
The first case considered is when the fragment shape in the fluoroscopic views exactly matches the preoperative plan.
We also considered the case of tracking a fragment with an unknown, or uncertain shape.
To achieve this we assumed that the nominal fragments represent preoperative plans for the randomly adjusted fragments present in the simulated fluoroscopic images.
The remaining two cases examined the effect of using intraoperative information to update the planned cut model.
For the first case of incorporating intraoperative knowledge, it was assumed that the ilium and pubis osteotomies would be perfectly recovered; the planned ilium and pubis cuts were replaced with the actual values.
In the second case, synthetic 2D cut annotations were used to simulate the fragment shape estimation described in section \ref{sec:method_frag_shape}.

Further details regarding the simulated datasets are described in supplementary section \ref{sec:supp_method_sim_study}.
\subsection{Cadaver Study} \label{sec:method_cadaver_study}
\begin{figure}[!t]
\centerline{\includegraphics[width=\columnwidth]{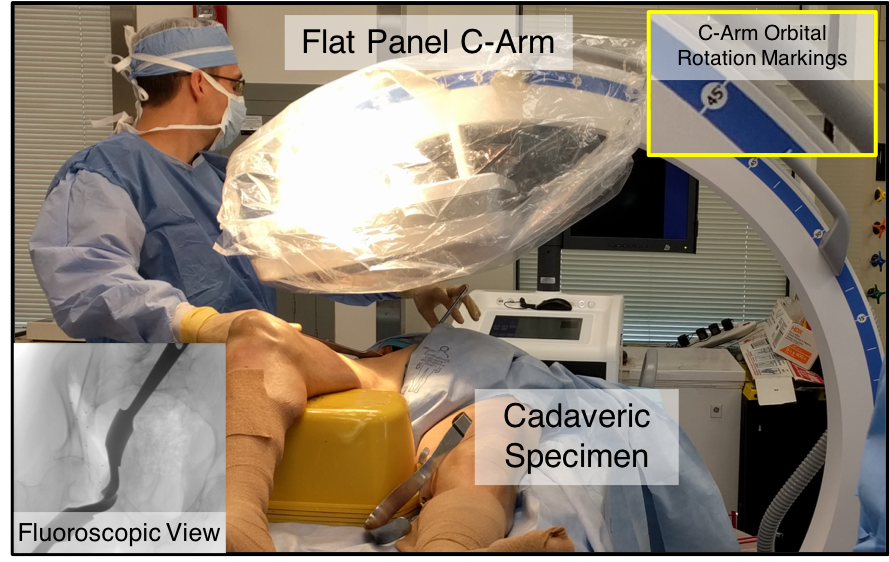}}
\caption{Intraoperative setup for the PAO procedure performed on the right side of a cadaveric specimen.
The current fluoroscopic view is shown in the bottom left.
The enlarged region in the top right shows the markings used to control the orbital rotation of the C-Arm.}
\label{fig:surgery_view}
\end{figure}
In order to further evaluate the proposed methods, intraoperative fluoroscopy was collected during six of the PAOs previously used for simulation studies.
Four specimens, three male and one female, were used.
Each surgery was performed under fluoroscopic guidance with a Siemens CIOS Fusion C-Arm with 30 cm flat panel detector.
Fig. \ref{fig:surgery_view} shows a photograph of the operating theater.
Each fluoroscopy image of the repositioned fragments contains screws, K-wires, or a combination of the two.

Fragment registrations were evaluated using the true fragment shape, preoperative planned fragments, and intraoperatively estimated fragment shapes.
Postoperative CT scans were taken, and the fragment shape was manually segmented, to obtain the true fragment shape.
Fifteen preoperative plans for each surgery were created by fitting cutting planes to the manually segmented fragment, followed by random adjustments.
Each preoperative plan was manually verified to be a valid PAO, with any invalid plan rejected and resampled.
The 2D cut lines of the ilium and pubis osteotomies were manually annotated in the fluoroscopic images, and were used to compute the ilium and pubis cutting planes of the estimated fragment shape.
Each estimated shape is completed with the planned ischium and posterior osteotomies.

Included with this paper is a supplementary video\footnote{https://youtu.be/pXmVa3-MJXo}, which illustrates the workflow used to perform the fragment shape and pose estimation for the left side of the specimen shown in Figures \ref{fig:frag_view_3d}, \ref{fig:frag_xrays}, \ref{fig:cut_2d_annotate}, and \ref{fig:surgery_view}.
This will be available at http://ieeexplore.ieee.org.

Ground truth poses of each fragment were obtained using metallic BBs implanted into the pelves.
The postoperative CT volumes contain the relocated fragment which had been intraoperatively fixed in place using screws and K-wire.
Prior to osteotomy, BBs were implanted onto the interior ilium surface in a region expected to lie on the fragment, and additional BBs were implanted onto the interior ilium surface in a region expected to not lie on the fragment.
The ground truth fragment poses are calculated with 3D/3D paired-point registrations \cite{horn1987closed} between the 3D locations of the fragment BBs before osteotomy, and the 3D locations after osteotomy.
For two specimens, and three PAOs, the fragment BB locations were computed using 3D CT imaging only.
Eight BBs were placed on the ilium region and another eight placed on the fragment, with each eight BB constellation made of four $1.5$ mm diameter BBs and four $1$ mm diameter BBs.
A Halifax Biomedical Inc. injection device was used to insert the $1$ mm diameter BBs.
One of our clinical co-authors, C.P.A., inserted these BBs after performing soft-tissue dissections.
An additional preoperative CT of the specimens was obtained to recover the pre-osteotomy BB locations.
For the remaining two specimens, and remaining three PAOs, four BBs ($1.5$ mm diameter) were inserted on the ilium region and four BBs ($1.5$ mm diameter) inserted on the fragment region.
No additional preoperative CT was obtained, however a series of fluoroscopic images were taken prior to osteotomy.
Using the BB constellations obtained from the postoperative CT, the pre and post-osteotomy 3D locations of the BBs were obtained through a series of 2D/3D landmark registrations on the fluoroscopic images.
A similar approach, using a plastic pelvis phantom with injected BBs and three views from CBCT projections, showed that the fragment pose may be recovered within $1\degree$/$1$ mm of CBCT ground truth \cite{armand2018biomechanical,murphy2013computer}.
Details of BB insertion and the methods used to obtain ground truth poses are described in supplementary section \sref{sec:supp_method_cadaver_study}.
The BBs were not present in the preoperative CTs used to drive the intensity-based registrations and did not aid the proposed methods in any way.
%
%
%
\section{Results}
\label{sec:results}
%
\subsection{Simulation Study}
\begin{figure*}[!t]
\centerline{\includegraphics[width=\textwidth]{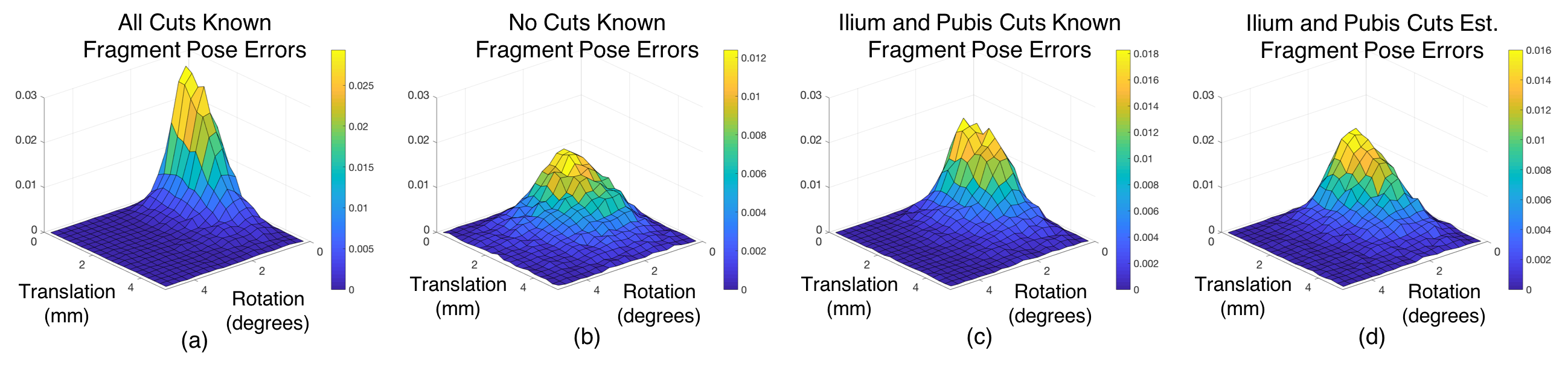}}
\caption{Normalized 2D histograms of fragment pose error for the simulation studies.
(a) The actual fragment shape is known and matches the planned fragment shape,
(b) the actual fragment shape is not known and none of the osteotomies match the planned fragment,
(c) the actual fragment is partially known, with the ilium and pubis osteotomies matching the planned fragment,
(d) the actual fragment is not known, but the ilium and pubis osteotomies are estimated from 2D cut lines.
}
\label{fig:sim_frag_hists}
\end{figure*}

Fragment pose estimation was most accurate when the exact fragment shape was used during registration, and accuracy monotonically decreased as knowledge of the shape decreased.
For each simulation in which the osteotomies were at least partially known or estimated, the mean rotation and translation errors were less than $2\degree$ and $2$ mm, respectively.
Mean fragment pose errors of $2.2\degree$ and $2.1$ mm were obtained from the simulation using preoperatively planned shapes.
For all fragment shapes, mean LCE errors were less than $1\degree$.
Table \ref{tab:sim_frag_pose_errors} lists the mean and standard deviations of the fragment pose and LCE errors for each simulation study.
Fig. \ref{fig:sim_frag_hists} shows the joint histograms for the fragment pose estimation errors.
The shape of the error distribution ``widens'' as the amount of uncertainty associated with the fragment shape increases, particularly with respect to rotation error.
The percentages of registrations with rotation errors below $3\degree$ were $96\%$,
$83\%$,
$92\%$,
and $91\%$,
for the cases of all cuts known, no cuts known (preoperatively planned), known ilium and pubis cuts, and estimated ilium and pubis cuts, respectively.
Similarly, the corresponding percentages of trials with LCE errors below $3\degree$ were $99\%$, $97\%$, $98\%$, and $98\%$.

Single-tailed Mann-Whitney U-Tests were performed to determine any statistical significance between the errors obtained when running registration with different fragment shapes.
The p-values are shown in Table \ref{tab:sim_pvals}, and indicate that pose and LCE estimation errors associated with the various categories of fragment shape are statistically different.
Moreover, the single-tailed test indicates that the errors associated with fragment shapes using less intraoperative osteotomy information are statistically larger than errors determined by fragment shapes incorporating more knowledge of the intraoperative cuts.
%
%
\begin{table*}[t]
\centering
\caption{Simulation Study Fragment Average Pose and Lateral Center Edge Angle Errors}
\setlength{\tabcolsep}{3pt}
\begin{tabular}{| p{72pt} |c|c|c|c|c|c|c|c|c|}
\hline
\multirow{2}{*}{Study}  &  \multicolumn{4}{c|}{Rotation ($\degree$)} &  \multicolumn{4}{c|}{Translation (mm)} & \multirow{2}{*}{{LCE ($\degree$)}}  \\ \cline{2-9}
& AP & LR & IS & Total & AP & LR & IS & Total &  \\
\hline
All Cuts Known            & $0.47 \pm 1.48$ & $0.73 \pm 1.35$ & $0.67 \pm 1.14$ & $1.25  \pm 2.24 $  & $1.54 \pm 1.29$ & $0.32 \pm 0.59$ & $0.43 \pm 0.59$ & $1.73 \pm 1.42$ &  $0.50 \pm 1.53$ \\ 
No Cuts Known            & $0.60 \pm 1.62$ & $1.53 \pm 1.77$ & $1.07 \pm 1.29$ & $2.19 \pm 2.58 $  & $1.84 \pm 1.46$  & $0.38 \pm 0.63$ & $0.50 \pm 0.58$ & $2.05 \pm 1.56$ & $0.88 \pm 1.72$ \\ 
Ilium \& Pubis Known   & $0.50 \pm 1.59$ & $1.07 \pm 1.64$ & $0.83 \pm 1.28$ & $1.63 \pm 2.54$   & $1.74 \pm 1.38$  & $0.33 \pm 0.61$ & $0.49 \pm 0.60$ & $1.94 \pm 1.49$ & $0.61 \pm 1.66$ \\
Ilium \& Pubis Est.        & $0.58 \pm 1.67$ & $1.16 \pm 1.78$ & $1.00 \pm 1.24$ & $1.86 \pm 2.62$  & $1.75 \pm 1.48$  & $0.34 \pm 0.57$ & $0.49 \pm 0.57$ & $1.96 \pm 1.56$  & $0.73 \pm 1.82$ \\ \hline
\multicolumn{10}{p{490pt}}{Means and standard deviations of the fragment pose errors and lateral center edge (LCE) angle errors for each simulation study.
The errors are organized by the type of fragment shape used for pose estimation, the total rotation and translation error magnitudes, the decompositions of the errors about each anatomical axis, and the LCE angle.
Small LCE errors for all fragment shapes indicate that any shape is reliable for reporting lateral coverage of the femoral head.}
\end{tabular}
\label{tab:sim_frag_pose_errors}
\end{table*}
%
%
\begin{table}[t]
\centering
\caption{Simulation Study Statistical Test Results}
\setlength{\tabcolsep}{3pt}
\begin{tabular}{| p{70.5pt} |  p{70.5pt} | p{33pt} | p{33pt} | p{33pt} |}
\hline
\multicolumn{2}{|c}{Fragment Shapes} & \multicolumn{3}{|c|}{p-Value}  \\ \hline
\multicolumn{1}{|c}{1}  &  \multicolumn{1}{|c|}{2}  &      \multicolumn{1}{c|}{Rot.}     &  \multicolumn{1}{c|}{Trans.}    &  \multicolumn{1}{c|}{LCE}  \\ \hline
All Cuts Known           &  No Cuts Known            &    \multicolumn{1}{c|}{$< 10^{-6}$}   &  \multicolumn{1}{c|}{$< 10^{-6}$}   & \multicolumn{1}{c|}{$< 10^{-6}$} \\
All Cuts Known           &  Ilium \& Pubis Known   &    \multicolumn{1}{c|}{$< 10^{-6}$}   &  \multicolumn{1}{c|}{$< 10^{-6}$}   & \multicolumn{1}{c|}{$< 10^{-6}$} \\
All Cuts Known           &  Ilium \& Pubis Est.        &    \multicolumn{1}{c|}{$< 10^{-6}$}   &  \multicolumn{1}{c|}{$< 10^{-6}$}   & \multicolumn{1}{c|}{$< 10^{-6}$}\\
Ilium \& Pubis Known &  Ilium \& Pubis Est.        &    \multicolumn{1}{c|}{$< 10^{-6}$}   &  \multicolumn{1}{c|}{$0.74$}           & \multicolumn{1}{c|}{$< 10^{-6}$}\\
Ilium \& Pubis Est.      & No Cuts Known             &    \multicolumn{1}{c|}{$< 10^{-6}$}   &  \multicolumn{1}{c|}{$< 10^{-6}$}   & \multicolumn{1}{c|}{$< 10^{-6}$} \\
 
\hline
\multicolumn{5}{p{245pt}}{Results of the Mann-Whitney U-Tests on fragment pose errors for the simulation studies.
										The total rotation and translation magnitudes of the pose errors, along with the error of the lateral center edge (LCE) angle are examined.
														 The null hypothesis indicates that both sets of errors are drawn from the same distribution with identical medians and
														 the alternative hypothesis indicates that the errors are drawn from different distributions with the second distribution (fragment shape with label 2) having larger median.
														 Applying a threshold of $0.005$ to the above p-values results in a rejection of the null-hypothesis for all but one of the tests.
														 This implies that errors associated with fragment shapes incorporating less information of the true cuts are statistically larger than the errors associated with cuts that more closely match intraoperative cuts,
														 except when comparing translation errors when ilium and pubis cuts are known versus estimated.}
\end{tabular}
\label{tab:sim_pvals}
\end{table}
%

\subsection{Cadaver Study}
A summary of the fragment pose and LCE error statistics is provided in Table \ref{tab:cadaver_pose_errors}.
Two results, one corresponding to a planned fragment shape and another to an estimated shape, produced fragment rotation errors above $20\degree$ and were discarded from analysis.
The pose estimations with manual segmented fragments, planned fragments, and estimated fragments resulted in mean rotation/translation errors of $2.2\degree$/$2.2$ mm, $3.8 \degree$/$2.5$ mm, and $3.5 \degree$/$2.5$ mm, respectively.
Mean LCE errors were below $2\degree$ for all fragment shapes.
Additionally, the standard deviations of all rotation and LCE errors strictly increased as knowledge of the fragment's 3D shape decreased.
Rotation errors were less than $3 \degree$ for $83\%$, $39\%$, and $39\%$ of the registration trials when using a manual segmentation of the fragment, a planned fragment, and an estimated fragment, respectively.
The corresponding proportions of LCE errors less than $3 \degree$ were $100\%$, $85\%$, and $85\%$.

Table \ref{tab:cad_pvals} shows p-values from statistical tests of significance conducted on the rotation, translation, and LCE errors.
The protocol was identical to that used for evaluating the simulation study results.
Using a threshold of $p=0.005$, no statistical differences are indicated between the rotation, translation, or LCE errors when comparing the various fragment shapes.

The single-landmark initialization strategy was used for four of the six cases, and multiple-landmark initialization was used for the remaining two.
In the two cases requiring multiple-landmark initialization, the initial views were taken with the detector parallel to the operating table and did not match a typical AP view due to the pelvic tilt of the specimens.

An expected runtime of $25.3$ seconds was obtained by re-running the estimated-fragment pose estimation pipeline 20 times for each surgery.
The expected time required to manually choose a single 2D landmark, required for registration initialization, was $2$ seconds; the expected time for manual annotation of 2D cut lines, required for fragment shape estimation, was $51.2$ seconds.

The ground truth fragment poses from each surgery, the error statistics for each individual surgery, violin plots of the error distributions, and a breakdown of timings across the individual components of the registration pipeline are provided in supplementary section \ref{sec:supp_results_cad_study}.
%
%
\begin{table*}[t]
\centering
\caption{Cadaver Experiment Fragment Pose and Lateral Center Edge Angle Errors}
\setlength{\tabcolsep}{3pt}
\begin{tabular}{| p{72pt} |c|c|c|c|c|c|c|c|c|}
\hline
\multirow{2}{*}{Study}  &  \multicolumn{4}{c|}{Rotation Errors ($\degree$)} &  \multicolumn{4}{c|}{Translation Error (mm)} & \multirow{2}{*}{{LCE ($\degree$)}}  \\ \cline{2-9}
& AP & LR & IS & Total & AP & LR & IS & Total & \\
\hline
Man. Seg.        & $0.51 \pm 0.25$  & $1.76 \pm 0.93$ & $1.02 \pm 0.51$  & $2.22 \pm 0.73$ & $1.22 \pm 0.50$ & $1.03 \pm 0.79$ & $1.29 \pm 1.01$ & $2.18 \pm 1.10$ & $1.08 \pm 0.52$  \\
Planned           & $1.17 \pm 1.10$  & $2.59 \pm 1.70$ & $1.89 \pm 1.48$ & $3.77 \pm 1.88$ & $1.49 \pm 1.40$ & $1.06 \pm 0.90$ & $1.16 \pm 0.99$ & $2.48 \pm 1.51$ &  $1.80 \pm 1.39$  \\
Estimated        & $0.98 \pm 0.80$  & $2.62 \pm 1.38$ & $1.55 \pm 1.21$ & $3.46 \pm 1.47$ & $0.98 \pm 0.83$ & $1.07 \pm 0.83$ & $1.15 \pm 0.90$ & $2.17 \pm 0.94$ &  $1.80 \pm 1.18$  \\ \hline
\multicolumn{10}{p{490pt}}{Means and standard deviations of fragment pose and lateral center edge angle (LCE) errors for the cadaver surgeries.
										For each type of fragment shape used during registration, the total rotation and translation error magnitudes are reported, along with the errors about each anatomical axis and the LCE angle error.
										 The manual segmented fragment shape, planned fragment shape, and estimated fragment shapes roughly correspond to the simulation cases of a fragment shape with all cuts known, a preoperatively planned fragment shape with no cuts known, and a shape with ilium and pubis cuts estimated, respectively, as shown in Table \ref{tab:sim_frag_pose_errors}.}
\end{tabular}
\label{tab:cadaver_pose_errors}
\end{table*}
%
%
%
\begin{table}[t]
\centering
\caption{Cadaver Experiments Statistical Test Results}
\setlength{\tabcolsep}{3pt}
\begin{tabular}{| p{76pt} |  p{76pt} | p{33pt} | p{33pt} | p{33pt} |}
\hline
\multicolumn{2}{|c}{Fragment Shapes} & \multicolumn{3}{|c|}{p-Value}  \\ \hline
\multicolumn{1}{|c}{1}  &  \multicolumn{1}{|c|}{2}  &      \multicolumn{1}{c|}{Rot.}     &  \multicolumn{1}{c|}{Trans.}    &  \multicolumn{1}{c|}{LCE}  \\ \hline
Man. Seg.   &  Planned                              &      \multicolumn{1}{c|}{$0.0194$}      &       \multicolumn{1}{c|}{$0.3798$}   &  \multicolumn{1}{c|}{$0.1194$}          \\
Man. Seg.   &  Ilium \& Pubis Cuts Est.      &      \multicolumn{1}{c|}{$0.0187$}     &        \multicolumn{1}{c|}{$0.5208$}   &  \multicolumn{1}{c|}{$0.0741$}       \\
Ilium \& Pubis Cuts Est. & Planned            &      \multicolumn{1}{c|}{$ 0.1678$}     &       \multicolumn{1}{c|}{$0.1023$}    &  \multicolumn{1}{c|}{$0.7167$}     \\
\hline
\multicolumn{5}{p{245pt}}{Results of the Mann-Whitney U-Tests on fragment pose and lateral center edge (LCE) angle errors for the cadaver experiments.
														See Table \ref{tab:sim_pvals} for a discussion of the hypotheses.
														Using a threshold of $0.005$, no significant difference is reported for any of the errors and for each comparison of shapes.}
\end{tabular}
\label{tab:cad_pvals}
\end{table}
%
\section{Discussion}
\label{sec:discussion}
The simulation study yielded promising results for the localization of the acetabular fragment.
As one would expect, the pose errors were smallest on average when the exact shape of each bone was known during registration, and degraded as less information about the osteotomies was known.
This increase in average error was consistent across total rotation and translation magnitudes, the decompositions about anatomical axes, and for the LCE angle estimates.
The increases in rotation error magnitudes and LCE angle errors were statistically significant for all comparisons and highlight the impact of preoperative/intraoperative fragment shape mismatch on registration accuracy.
Increases in the translation error magnitudes were statistically significant in all but one comparison: using true pubis and ilium osteotomies, mixed with planned ischial and posterior osteotomies, compared with estimating the pubis and ilium osteotomies.
This also indicates that the fragment localization is partially robust to differences between the planned and intraoperative osteotomies for the ischial and posterior cuts.

To our knowledge, no bounds on the accuracy of fragment pose or LCE angle have been identified in previous literature, therefore we consider $3\degree$ to be an acceptable upper bound on LCE angle error.
This threshold allows the clinician to target LCE angles between $33\degree-37\degree$, while having confidence that the true LCE angle is within the $30\degree-40\degree$ criteria put forth by \cite{hartig2012factors}.
At this threshold, the lowest success rate of $97\%$ corresponds to a preoperatively planned fragment shape, and indicates that an arbitrary preoperative plan is sufficient to obtain a successful registration.
However, if the LCE error threshold is lowered to $2\degree$, successful pose estimates were found in $98\%$, $92\%$, $97\%$, and $96\%$ of trials, for the cases of the fragment shape with all cuts known, the preoperatively planned fragment shape, the shape with known ilium and pubis cuts, and the shape with estimated ilium and pubis cuts, respectively.
Therefore, a clinician may attach more confidence to pose estimates obtained after estimating the fragment's ilium and pubis cuts.

In cadaver surgeries, registrations using each fragment shape yielded average LCE errors less than $2\degree$.
This indicates that the proposed pipeline is capable of producing clinically usable estimates on real fluoroscopic imagery.
In terms of average rotational and LCE error, registrations using the manually segmented fragment shape outperformed registrations using preoperatively planned shapes and intraoperatively refined shapes.
However, when examining translation dimensions, the manually segmented fragment did not offer a noticeable advantage over the remaining two shapes.
The performance across all error metrics was roughly equivalent when comparing the registrations using the preoperative plan and estimated fragments.
Moreover, when comparing the errors produced through registrations using differing shape types, no statistically significant differences were found.
This is in stark contrast to the simulation study results, which imply that an estimated shape should perform statistically better than a preoperatively planned fragment.

We believe this is primarily caused by low bone qualities from two specimens used for three surgeries.
Prior to any osteotomies made to free the acetabular fragment, a small osteotomy of the anterior superior iliac spine (ASIS) is performed in order to provide sufficient access to the ilium.
At the conclusion of the surgery, the ASIS is reattached in its original location.
The ASIS osteotomy was excessively large for two specimens, resulting in a fragment shape with an additional osteotomy not representable by our four-cut model.
When these three surgeries are removed from the results, $100\%$ of the registrations using manually segmented fragment shapes, $95\%$ using preoperatively planned fragment shapes, and $100\%$ using estimated shapes produce LCE errors less than $3\degree$.
Furthermore, mean rotation errors of the fragment pose were $2.4\degree$, $4.0\degree$, and $3.1\degree$ when using a manually segmented fragment shape, a preoperative plan, and estimated shape, respectively.
Although no statistical difference was found between these error distributions, the results are more consistent with the trend established by the simulation study of increasing errors, corresponding to registrations using more mismatched fragment shapes.
Since most patients undergoing PAO are young adults, bone quality is not expected to be an issue in clinical use.
Considering these factors, we can say that the fragment pose estimation using an estimated shape provided clinically relevant results to determine femoral head coverage, with LCE errors less than $3\degree$.

The mean rotation error of $3.5\degree$, when using the estimated fragment shape during the cadaver surgeries, is noticeably larger than the $1.8\degree$ error reported by a state-of-the-art optically tracked solution \cite{murphy2015development}.
Translation errors found with the proposed system are roughly equivalent to the $2.2$ mm identified by the optically tracked method.
Although the X-Ray navigation approach has inferior rotation performance compared to the optically tracked method, the previous analysis of LCE angle errors indicate it is capable of providing clinically appropriate pose estimates.

The amount of femoral head coverage is primarily derived from the series of rotations applied to the acetabular fragment about the femoral head.
For this reason, we report errors using rotation and translation components of the pose differences between registered estimates and ground truth poses.
Furthermore, large TREs at regions distant from the cartilage surface, such as the ilium osteotomy, do not imply an incorrect estimate of the femoral head coverage.

Given the promising results in cadaveric experiments, and the diversity of data used in simulation, we believe this method should perform well on human subjects.
Although we did not evaluate the proposed methods on dysplastic hips, our method should not suffer any performance degradation when run on dysplastic hips of younger patients.
We have not made any assumptions in our processing specific to normal hip anatomy, and we only require an intensity volume of the pelvis that we may compute DRRs from.
Some dysplastic hips may be subject to severe arthritis, which would likely cause difficulty during preoperative segmentation and also create a less-smooth registration similarity metric,
however these patients would most likely undergo a total hip replacement (THR) instead of pelvic osteotomy.

The expected computation time of approximately $25$ seconds is reasonable for intraoperative operation, however this does not include the time required to perform any manual annotations.
Since the fragment will not change shape after the osteotomies are completed, annotation of the cut lines only needs to be performed once.
It is likely that several adjustments to the fragment's pose will be required, therefore we believe that the cut annotation time is amortized over each of these adjustments.
A similar amortization is not applicable to registration re-initialization via landmark identification.
However, other techniques using neural networks for recognizing fluoroscopic landmarks \cite{bier2018x}, or for approximate pose regression \cite{kendall2015posenet}, could be applied to avoid this bottleneck.


Although implicitly penalized by the image similarity metric, the proposed registration method does not penalize 3D collision or overlap of bones.
It is possible that the method's performance may improve by including collision and overlap information into the regularization term of \eqref{eq:intensity_regi_opt_prob}.
When the fragment shape used during registration does not match the true fragment shape, it is possible for collisions to occur, particularly about the ischial and posterior osteotomies.
Therefore, it may be possible to recover more accurate shapes by updating the volumetric segmentation to remove collisions after registration.
Shapes could be refined by iteratively switching between registration and shape updates.

The six cadaver surgeries have served as proof-of-concept for the methods proposed in this work, however we believe further efforts should include clinical PAO cases.
This would include the acquisition of clinical PAO fluoroscopic data used for further validation, and move towards a comparison of patient outcomes and operative times of non-navigated, optically navigated, and X-Ray navigated cases.
The system could also integrate a preoperative plan of the biomechanically optimal fragment reposition, and intraoperatively provide feedback consisting of adjustments that would maximize the likelihood of achieving the plan.

Although the experiments in this paper were restricted to PAO, we have not made any assumptions that would preclude application to other pelvic osteotomies,
such as RAO, CPO, or triple osteotomy.
Forty percent of THR surgeries result in a mal-positioned (with respect to the ``safe zone'') acetabular component \cite{callanan2011john}, therefore we believe that our fragment tracking approach would also be useful for the intraoperative visualization, and guidance, of the THR acetabular implant.

Although the use of a preoperative CT scan is an increasingly common practice, it is not considered the current standard of care at all centers.
When CT scans are used for diagnosis or preoperative planning, slice spacings are typically in the range of $2-3$ mm \cite{troelsen2010assessment}.
It is not clear what effect lower resolution CT data will have on the accuracy of the proposed system and remains an important topic to be investigated.
Furthermore, higher resolution preoperative CT scans may add additional cost and expose the patient to, potentially hazardous, radiation.
In the future, a ``CT free'' approach could be implemented using a statistical deformation model (SDM) created from a database of existing pelvic CTs \cite{otake2015comparison}.
The SDM would be used in a deformable 2D/3D registration to recover the patient's 3D anatomy from standard preoperative standing radiographs and several pre-osteotomy fluoroscopic images \cite{sadowsky2007deformable}.
However, the SDM would most likely be unable to recover a sufficiently accurate cartilage model required for biomechanical measures and analysis.
This could be overcome through the use of a partial preoperative CT, scanning only the acetabulum and extrapolating the remaining anatomy with the SDM, similar to the methods proposed in \cite{grupp2016pelvis} and \cite{chintalapani2010statistical}.
%
\section{Conclusion}
\label{sec:conclusion}
The proposed method for intraoperative localization of a mobilized acetabular fragment, with uncertain shape, has been demonstrated to report LCE angles within clinical tolerance in simulated cases and also in cadaveric cases for PAO.
Less common equipment used in other approaches, such as an encoded or motorized CBCT C-Arm or an optical tracking device, is not required by this method.
Additionally, the processing runs in an amount of time which is reasonable for incorporation into an intraoperative workflow.
As entry-level flat panel imaging technology becomes more common in operating theaters, our method should be directly applicable to pelvic osteotomies or other similar fragment tracking tasks.
The proposed X-Ray navigation could make pelvic osteotomies more accessible to novice surgeons, reduce the number of times a fragment needs to be adjusted during a procedure, and possibly reduce the radiation exposure to the surgical team by reducing the number of fluoroscopic images used.

\section*{Acknowledgment}

The authors would like to thank Mr. Demetries Boston for his assistance during cadaveric testing.
Professor Yoshinobu Sato provided productive and insightful discourse during the algorithmic development of this work, for which we are very grateful.
Additionally, we are indebted to Dr. Masaki Takao and Dr. Nobuhiko Sugano for allowing us to observe an RAO procedure and become aware of the challenges involved.
We also thank Professor Mathias Unberath for helpful discussions.
Finally, we are appreciative of the constructive comments and critiques provided by the anonymous reviewers, which have resulted in an improved paper.
This feedback led to the single-landmark registration initialization strategy, and has greatly improved the system's ease of use.
This research was supported by NIH/NIBIB grants R01EB006839, R21EB020113,
MEXT/JSPS KAKENHI 26108004,
Johns Hopkins University Internal Funds, 
and a Johns Hopkins University Applied Physics Laboratory Graduate Student Fellowship.
Part of this research project was conducted using computational resources at the Maryland Advanced Research Computing Center (MARCC).

\bibliographystyle{IEEEtran}
\bibliography{IEEEabrv,refs} 
\appendix[Supplementary Material]
%
\renewcommand{\thetable}{S-\arabic{table}}
\renewcommand{\thefigure}{S-\arabic{figure}}
\setcounter{figure}{0} 
\setcounter{table}{0} 
%
\section{Materials and Methods} \label{sec:supp_methods}
%
\subsection{Preoperative Processing and Planning}\label{sec:supp_preop_proc_plan}
Preoperative CT scans of the full pelvis and proximal femur for each cadaveric specimen were measured with a Toshiba Aquilion ONE.
Table \ref{tab:supp_preop_ct_params} shows pixel spacings, slice spacings, and slice thicknesses for each scan.

After resampling to $1$ mm isostropic voxel spacings, the pelvis and femurs of each specimen were segmented using an implementation of \cite{krvcah2011fully}.
The average runtime of this method was $50.8 \pm 13.0$ seconds when run on a Late 2013 MacBook Pro with a 2.3 GHz Intel Core i7-4850HQ CPU and 16 GB of RAM.
For three of the joints, a manual touch was required to ensure proper separation of the acetabulum and femoral head; this took an average of $16.0 \pm 4.6$ minutes.
Since this touch up was performed with a generic segmentation module in 3D Slicer \cite{fedorov20123d}, it is possible that the touch up could be accelerated with software tailored to this task, or even eliminated through the use of more advanced automatic pelvis segmentation methods \cite{chu2015facts,chu2015mascg,yokota2013automated}.

\begin{figure}[t]
\centerline{\includegraphics[width=\columnwidth]{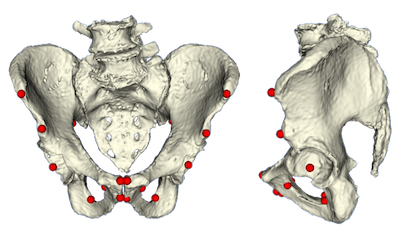}}
\caption{An example of 3D landmarks manually identified in a preoperative CT for establishment of the anterior pelvic plane coordinate frame, and for initialization of the registration pipeline.
The specimen shown here also corresponds to Figures \ref{fig:frag_view_3d}, \ref{fig:frag_xrays}, \ref{fig:cut_2d_annotate}, and \ref{fig:surgery_view}.}
\label{fig:supp_landmarks_3d_viz}
\end{figure}

The manual selection of landmarks used for establishing the anterior pelvic plane (APP) coordinate frame, and possible use in registration initialization, took an average of $132.7 \pm 7.1$ seconds using 3D Slicer.
An example of these landmarks is shown in Fig. \ref{fig:supp_landmarks_3d_viz}, and consists of bilateral annotations of the anterior superior iliac spine (ASIS), anterior inferior iliac spine (AIIS), center of femoral head (FH), superior pubic symphysis (SPS), inferior pubic symphysis (IPS), medial obturator foramen (MOF), inferior obturator foramen (IOF), and the greater sciatic notch (GSN).
The ipsilateral MOF is used for all single-landmark registration initializations.

The average time to annotate points along the trace of the acetabular rim was $70.9 \pm 17.1$ seconds.
These points are required in order to produce intraoperative estimates of the lateral center edge angle.
%
%
%
\begin{table}[t]
\centering
\caption{CT Volume Pixel Spacings and Slice Thicknesses}
\setlength{\tabcolsep}{3pt}
\begin{tabular}{| p{32pt} | p{60pt} | p{60pt} | p{60pt} |}
\hline
Specimen & In-Plane Spacing (mm/pixel) & Out-of-Plane Spacing (mm) & Slice Thickness (mm) \\
\hline
1* & $0.740$ & $0.5$ & $0.5$ \\
2* & $0.896$ & $0.5$ & $0.5$ \\
3 & $0.839$ & $0.5$ & $0.5$ \\
4 & $0.850$ & $0.5$ & $0.5$ \\
5* & $0.881$ & $0.5$ & $0.5$ \\
6* & $0.873$ & $0.5$ & $0.5$ \\
\hline
\multicolumn{4}{p{245pt}}{Isotropic in-plane pixel spacings, constant out-of-plane slice spacings, and slice thicknesses for each of the preoperative cadaver CT scans.
All specimens were used in the simulation study, and an * indicates that the specimen was used as part of the cadaver study.}
\end{tabular}
\label{tab:supp_preop_ct_params}
\end{table}
%
%
%
\subsection{Intraoperative Registration}\label{sec:supp_gen_regi}
When it is not possible to use the single-landmark initialization strategy, the system is initialized using at least four landmarks.
When identifiable in the 2D image, each of the preoperatively defined 3D registration landmarks, $\mathbf{p}_\text{3D}^{(i)}$, is manually annotated in the 2D image, $\mathbf{p}_\text{2D}^{(i)}$.
An initial pose is computed with the POSIT method \cite{dementhon1995model}, which is then used as the initialization to a nonlinear optimization of landmark re-projection error, described in \eqref{eq:landmark_2d3d}.
The solution of \eqref{eq:landmark_2d3d} is used to initialize the intensity-based registration of the initial view.
\begin{equation} \label{eq:landmark_2d3d}
	\min_{\theta \in SE(3)} \sum_i \frac{1}{2} \left\lVert \mathbf{p}^{(i)}_{\text{2D}} - \mathcal{P}(\mathbf{p}^{(i)}_{\text{3D}}; \theta) \right\rVert_2^2
\end{equation}
POSIT works well to initialize this process, since it requires no initial pose estimate as input and the scaled orthographic assumption is reasonably accurate, given that the C-Arm has a large focal length and the anatomical landmarks are relatively close to each other with respect to depth.

The center of orbital rotation and the rotation axis of the CIOS Fusion was not provided by the manufacturer, therefore an offline calibration process similar to \cite{navab19983d} was conducted.
Using the single-view registration strategy, previously described in section \ref{sec:gen_regi}, a sawbones pelvis was used as a fiducial to perform 2D/3D registrations at 41 C-Arm poses along the main orbit.
From these registrations, the isocenter and main orbital axis of rotation were computed.

CMA-ES population and initial sigma parameters for each registration phase are shown in Table \ref{tab:regi_cmaes_param}.

\begin{figure}[t]
\centerline{\includegraphics[width=\columnwidth]{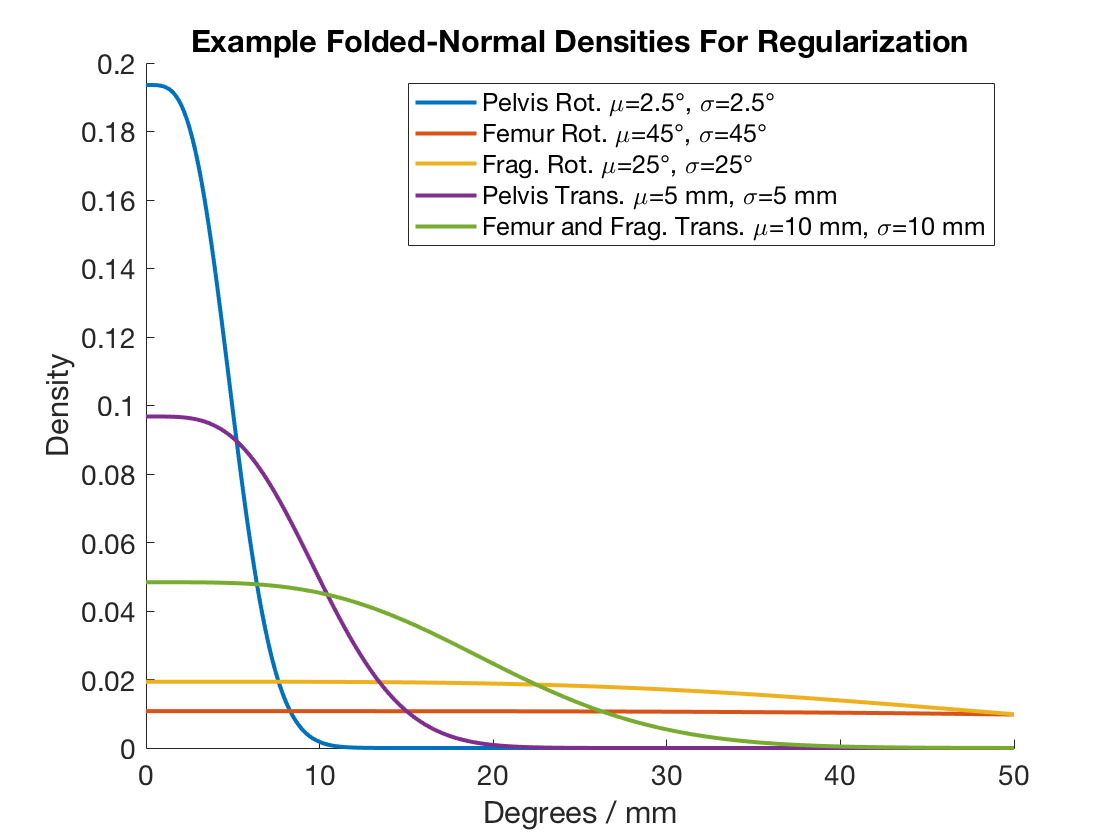}}
\caption{Plots of the Folded-Normal densities used for regularization in the multiple-object/multiple-view registrations.
Each density used has the two parameters set equal ($\mu = \sigma$).
This creates a flatter curve around zero, which smoothly decreases asymptotically to zero.
This behavior simulates a soft box-constraint, so that poses about the initialization are uniformly likely, and slowly become less likely for values approximately greater than $\mu$.}
\label{fig:fold_norm_dists}
\end{figure}

The regularization functions applied during the CMA-ES optimization phases are modeled as the log-probability densities of pose parameters.
For single view registrations of the pelvis, each translation component and Euler angle are assumed to be drawn from independent single-variate Normal distributions.
The exact parameters of these distributions are shown in Table \ref{tab:single_view_reg_params}.
Folded-Normal distributions are used to model the probability densities of the total rotation and translation magnitudes of each object's pose in the multiple-object, multiple-view, registrations.
Parameter values and plots of the Folded-Normal distributions used in this paper are shown in Fig. \ref{fig:fold_norm_dists}.
When combining the image similarity score and regularization value as a weighted sum, the image similarity term is given a weight of $0.9$ and the regularization term is weighted at $0.1$.

BOBYQA box constraints used in each registration phase are shown in Table \ref{tab:bobyqa_box_constraints}.

The diameter of the patch used for similarity metric computations is defined to have a diameter of 83 pixels in the full resolution 2D images, and is scaled at each resolution level proportionally to the downsample ratio.
%
%
%
\begin{table}[t]
\centering
\caption{CMA-ES Parameters}
\setlength{\tabcolsep}{3pt}
\begin{tabular}{|c|c|c|c|}
\hline
View & Init./Object & Pop. Size & $\sigma$ $(\degree,\degree,\degree,\text{mm},\text{mm},\text{mm})$ \\
\hline
\multirow{3}{*}{Single-View Pelvis} & Landmark Init. & $100$ & $15,15,30,50,50,100$ \\
                                             & Rotation Init. & $20$ & $2.5,2.5,5,2.5,2.5,2.5$ \\
                                             \hline
\multirow{3}{*}{Multi-View} & Pelvis & $20$ & $2.3,2.3,2.3,2,2,2$ \\
                                           & Femur & $100$ & $17.2,17.2,17.2,5,5,5$ \\
                                           & Fragment & $100$ & $17.2,17.2,17.2,5,5,5$ \\
\hline
\multicolumn{4}{p{245pt}}{The single-view $\sigma$ parameters are referenced at the in the camera frame, and the multi-view parameters are referenced in the APP frame.}
\end{tabular}
\label{tab:regi_cmaes_param}
\end{table}
\begin{table}[t]
\centering
\caption{Single-View Regularization Parameters}
\setlength{\tabcolsep}{3pt}
\begin{tabular}{| p{80pt} | p{20pt} | p{20pt} | p{20pt} | p{20pt} | p{20pt} | p{20pt} |}
\hline
\multirow{2}{*}{Initialization}  &  \multicolumn{3}{c|}{Rotation ($\degree$)} &  \multicolumn{3}{c|}{Translation (mm)}  \\ \cline{2-7}
& $\sigma_X$ &   $\sigma_Y$ & $\sigma_Z$ & $\sigma_X$ & $\sigma_Y$ & $\sigma_Z$  \\
\hline
Landmark Init. & $10$ & $10$ & $10$ & $20$ & $20$ & $100$ \\ \hline
Rotation Init. & $2.5$ & $2.5$ & $2.5$ & $2.5$ & $2.5$ & $25$ \\
\hline
\multicolumn{7}{p{245pt}}{Standard deviation parameter values for the regularizations used for single-view registrations.
All distributions are zero-mean.
$X$ and $Y$ represent the in-plane directions and the $Z$ indicates the out-of-plane, or depth direction.}
\end{tabular}
\label{tab:single_view_reg_params}
\end{table}
%
\begin{table}[t]
\centering
\caption{BOBYQA Box Constraints}
\setlength{\tabcolsep}{3pt}
\begin{tabular}{|c|c|c|}
\hline
Registration Type & Object & Bounds $\pm (\degree,\degree,\degree,\text{mm},\text{mm},\text{mm})$ \\
\hline
Single View Land. Init. & Pelvis & $7.5,7.5,15,25,25,50$ \\
 \cline{1-3}
 Single View Rot. Init. & Pelvis & $1.25,1.25,1.25,1.25,1.25,12.5$ \\
 \cline{1-3}
\multirow{3}{*}{Multi-Obj./View Seq.} & Pelvis & $2.5,2.5,2.5,5,5,5$ \\
                          & Femur & $10,10,10,30,30,30$ \\
                         & Fragment & $10,10,10,30,30,30$ \\
                                             \cline{1-3}
\multirow{3}{*}{Multi-Obj./View Simult.} & Pelvis &  $2.5,2.5,2.5, 5,5,5$ \\
                          & Femur &  $2.5,2.5,2.5, 10,10,10$ \\
                           & Fragment &  $2.5,2.5,2.5, 10,10,10$ \\
                                             \hline
\multicolumn{3}{p{245pt}}{The constraints are applied to $\mathfrak{se}(3)$ elements, referenced at the initial pose estimate. The axes are aligned with the camera frame for single view registrations and the APP frame for multiple-object/multiple-view registrations.}
\end{tabular}
\label{tab:bobyqa_box_constraints}
\end{table}
%
%
\subsection{Fragment Shape Estimation} \label{sec:supp_method_frag_shape}
\begin{figure}[t]
\centerline{\includegraphics[width=\columnwidth]{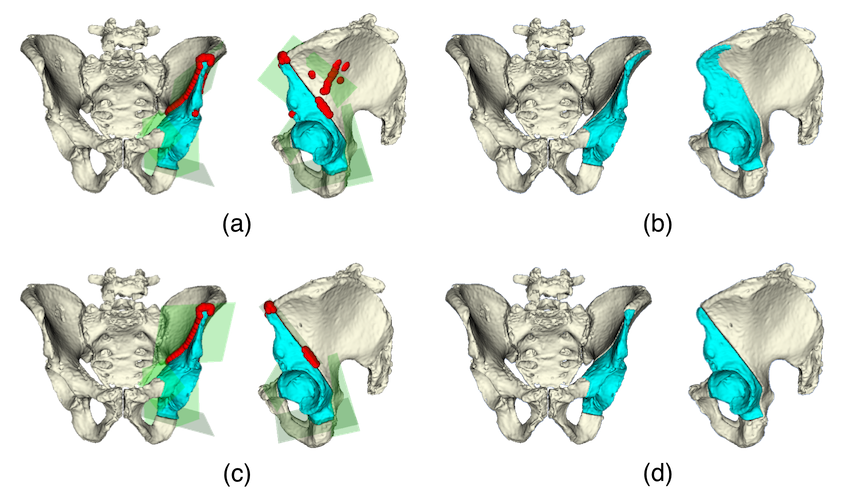}}
\caption{Examples of fragment shape estimation without outlier rejection (a),(b), and with outlier rejection (c),(d) applied on the reconstructed 3D cut points.
Figures (a) and (c) overlay the reconstructed 3D ilium cut points over the ground truth fragment shape; the cutting planes fit to these points are shown in green.
An implausible cutting plane is caused by the many outliers present in (a) due to the view geometry along, noisy 2D cut labels, and an imperfect registration.
By using the anatomy-based and RANSAC outlier removal process described in section \ref{sec:supp_method_frag_shape}, these outliers are not present in (c) and the ilium cutting plane intersects the ground truth cut.
The cutting planes shown in (a) generate the invalid fragment shape in (b), while the cutting planes shown in (c) generate the reasonable fragment shape shown in (d).
This example is taken from one of the simulated cases on the left side of specimen 4.}
\label{fig:cut_fitting_outlier_example}
\end{figure}
The estimated set of 3D osteotomy points often contains outliers as a result of human error when labeling osteotomy cut lines in 2D and pelvis registration errors in 3D.
Including outliers into the cutting plane estimation may produce unrealistic fragment shapes and yield poor registration performance.
We first perform an anatomy-based pruning of points, by only considering points on the ipsilateral side, and points above the femoral head for the ilium cut, and points in front of the femoral head for the pubis cut.
A RANSAC \cite{fischler1981random} strategy is then employed on the pruned set of points to obtain the final fit.
Fig. \ref{fig:cut_fitting_outlier_example} shows an unrealistic fragment created without any outlier removal and a corresponding realistic fragment after accounting for outliers.
%
\subsection{Simulation Study} \label{sec:supp_method_sim_study}
\begin{table}[t]
\centering
\caption{Simulation Study Fragment and Femur Sampling Parameters}
\setlength{\tabcolsep}{3pt}
\begin{tabular}{|c|c|c|c|}
\hline
Axis & Fragment Rotation (\degree) & Femur Rotation (\degree) & Translation (mm) \\
\hline
LR & $U[0,20]$ & $U[0,20]$ & $U[-2.5,7.5]$ \\
IS  & $U[0,6]$ & $U[-5,5]$ & $U[-4,0]$ \\
AP & $U[0,25]$ & $U[-2,2]$ & $U[-1,5]$ \\
\hline
\multicolumn{4}{p{245pt}}{The LR rotation always rotates the fragment ``forwards,'' and the signs for rotations about IS and AP and translation along LR are set depending on the patient side to return realistic movements.}
\end{tabular}
\label{tab:supp_frag_sampling}
\end{table}

The parameters used for sampling random fragment and femur poses are shown in Table \ref{tab:supp_frag_sampling}.

Soft-tissues were incorporated into our simulated fluoroscopic data by transforming the fragment and femur voxels in the original CT volume.
The warped fragment and femur intensity values overwrite any existing intensities.
Pelvis and contralateral femur intensities remain unchanged, since pose adjustments were sampled free of collision with boney structures.
Random HU intensities roughly corresponding to muscle tissue are used for any voxel that no longer corresponds to the fragment or femur.
The muscle intensities are randomly drawn from $N(\alpha,20)$ for each voxel location; $\alpha$ is drawn from $U[35,55]$ for each pair of fragment and femur relocations.

Two randomly configured K-wires are inserted into the volume at random, but anatomically relevant, poses.
The orientation of each wire will be determined by the choice of insertion points into the pelvis and the ending point on the acetabular fragment.
Each insertion point is uniformly sampled from the post-osteotomy iliac crest surface.
If the second wire's insertion point lies within $20$ mm of the first insertion point, it is rejected and resampled.
Denote the valid insertion point as $p_i$, let $n_i$ be the surface unit normal at $p_i$, and let $n_f$ be the unit vector oriented from $p_i$ to its closest point on the relocated acetabular fragment.
Define $n_{if}$ as the unit vector obtained from $\frac{1}{2} (n_i + n_f)$.
The direction defined by $n_{if}$ imposes a constraint that the wire be oriented towards the fragment, yet also be influenced by the curvature of the iliac crest.
Despite these constraints, it is possible for $n_{if}$ to yield unrealistic wire poses directed at parts of the fragment closer to the iliac spine, or to not intersect the fragment at all.
In order to overcome this, the normal is re-oriented towards the femoral head by combing $n_{if}$ with $n_{fh}$, the unit normal directed from $p_i$ to the femoral head, as shown in \eqref{eq:supp_kwire_normal}.
\begin{equation} \label{eq:supp_kwire_normal}
	\hat{n} = \frac{ \beta n_{fh} + \left( 1 - \beta \right) n_{if} }{ \left \| \beta n_{fh} + \left( 1 - \beta \right) n_{if} \right \|_2 }
\end{equation}
For a given value of $\beta \in [0,1]$, let $\hat{p}_f$ denote the intersection point of the ray, starting at $p_i$ and oriented by $\hat{n}$, with the relocated acetabular fragment.
The direction, $\hat{n}$, is considered valid when there is an intersection and $\| p_i - \hat{p}_f \|_2 \in [40, 110]$.
Values of $\beta$ are evaluated from $\{ 0.5, 0.6, 0.7, 0.8, 0.9, 1.0 \}$, in increasing order, until a valid $\hat{n}$ is computed.
If a valid $\hat{n}$ is not obtained, the process is restarted by sampling a new insertion point.
When a valid $\hat{n}$ is obtained, the K-wire's orientation is defined by $p_i$ and $\hat{p}_f$.
Two geometric primitives, a cylindrical body and a conical tip, define the shape of each wire.
One cap of the cylindrical body is located at $\hat{p}_f$, with the other cap oriented towards $p_i$ at a distance sampled from $U[190.5, 266.7]$ (mm).
Likewise, the bottom of the cone is coincident with the body at $\hat{p}_f$, with the tip oriented away from $p_i$, into the fragment body, with a height sampled from $U[4,6]$ (mm).
The radius of the body is sampled from $U[0.5, 1.5]$ (mm).
Each wire's attenuation is determined by a constant HU value sampled from $U[14000, 26000]$.
The wire is inserted into the volume by checking to see if each voxel is contained within the geometric primitives, and replacing the existing voxel's intensity with the wire's HU value.
In order to avoid aliasing, the volume is upsampled in each dimension by $4\times$ prior to K-wire insertion; the volume is downsampled back to its original resolution after the intensities are updated.

The updated HU volume is mapped into a volume of linear attenuation coefficients.
Similar to \cite{markelj2010standardized}, the HU conversion transform was shifted to be piecewise linear, with all HU values less than $-130$ mapping to zero attenuation, and then linearly progressing from zero for values greater than $-130$ HU.
Clamping and shifting in this manner attempts to account for the difference in X-Ray energies between the CT images and intraoperative fluoroscopic images.
No upper bound was used for clamping.

%
The simulated fluoroscopic image intensities are sampled from a Poisson distribution defined at each pixel location $(x,y)$.
Each pixel's distribution uses a mean parameter, $N(x,y)$, defined by the Beer-Lambert law shown in \eqref{eq:supp_bl_law}.
\begin{equation} \label{eq:supp_bl_law}
N \left(x,y \right) = N_0 \textrm{exp} \left \{ - \int_0^1 \mu \left( \ell \left( x, y, t \right) \right) dt \right \}
\end{equation}
The line connecting the X-Ray source to the detector location at $(x,y)$ is denoted as $\ell$, the volume of attenuation coefficients is written as $\mu$, and $N_0$ approximates the number of photons emitted from the X-Ray source.
$N_0$ was set to $2000$ for all experiments.

An initial AP view of the patient was simulated by first aligning the APP with the projective coordinate frame: matching the left-right (LR) and inferior-superior (IS) axes with the in-plane detector directions, and the anterior-posterior axis with the source-to-detector axis.
The patient was oriented supine with the detector located anterior of the patient, and with the center of the ipsilateral femoral head at $80\%$ of the source to detector distance and projecting to the center of the detector.
In order to include more of the pelvis in the field of view, the patient was translated $25$ mm along the LR direction and $35$ mm along the IS direction.
A diverse set of initial views was obtained by applying a random rigid transform in the APP frame.
Random rotations were sampled by first uniformly sampling a random rotation axis and then drawing a rotation angle from $U[-10\degree, 10 \degree]$.
Translation was sampled by uniformly sampling a direction, followed by a magnitude from $U[0,10]$ (mm).

Two additional views were obtained by sampling random orbital rotation angles, followed by small random perturbations.
Orbital rotation angles were sampled from $N(-10\degree, 3\degree)$ and $N(15\degree ,3 \degree)$.
Random perturbations were sampled analogously to the random rigid movement applied to AP views, except with rotation angles drawn from $U[-2\degree,2\degree]$ and translation magnitudes drawn from $U[-2,2]$ (mm).

%
\begin{figure}[t]
\centerline{\includegraphics[width=\columnwidth]{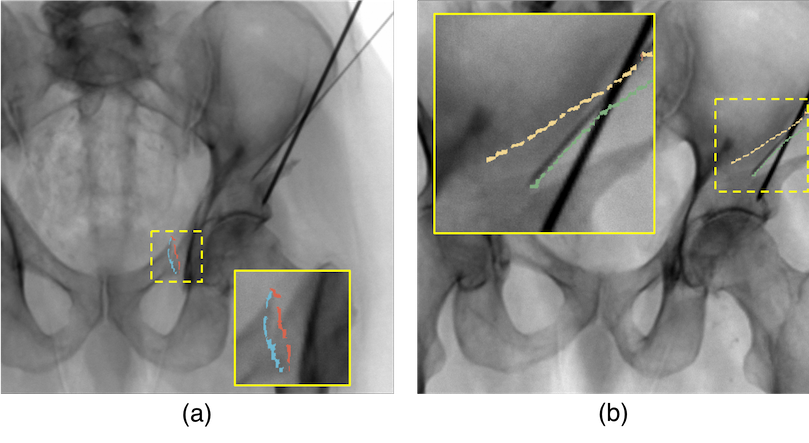}}
\caption{An example of the simulated 2D cut annotations.
Pubis cut annotations are shown in (a), with beige labels indicating rays entering the pelvis surface and green labels indicating rays leaving the pelvis surface.
Ilium cut annotations are shown in (b), with red labels indicating rays entering the pelvis surface and blue labels indicating rays leaving the pelvis surface.
Rays are cast oriented from the detector to the source.}
\label{fig:supp_sim_2d_cut_annotations}
\end{figure}
Initial 2D ilium and pubis simulated cut annotations were computed by ray casting and checking for collision with the ground truth intersection of each cut with the pelvis surface.
When annotating cuts in the cadaver fluoroscopy images, it was typical to have discontinuities in the annotation due to a lack of contrast (Fig. \ref{fig:cut_2d_annotate}c);
this was mimicked by moving along the principal axis of each cut line, and randomly erasing groups of labels.
The cut labels were randomly warped to simulate human error in the annotation process.
At each corner of the 2D cut line oriented bounding box, random deformations were sampled, and a deformation vector was computed at each pixel within the oriented bounding box using linear interpolation.
The simulated 2D cut annotations, along with full pelvis, single-view, registrations were used as input to the shape estimation method described in section \ref{sec:method_frag_shape}.
Fig. \ref{fig:supp_sim_2d_cut_annotations} shows an example of the simulated annotations.
%
\subsection{Cadaver Study} \label{sec:supp_method_cadaver_study}
As introduced in section \ref{sec:method_cadaver_study}, ground truth estimates of the fragment poses were obtained using metallic BBs inserted into the surface of the pelvis.
All $1.5$ diameter BBs were put in place by creating a small bone burr into the pelvis surface, applying cyanoacrylate (``Super Glue'') to the burr, inserting the BB into the burr, and then reapplying the cyanoacrylate.
The first method used for ground truth pose uses a combination of $1.5$ mm and $1$ mm BBs.
A Halifax Biomedical Inc.\footnote{http://halifaxbiomedical.com/products/halifax-bead-inserter} injection device was used to insert the $1$ mm diameter BBs.
Using a mallet, some force was applied to the injector in order to pierce the cortical wall about the acetabular region.
The second ground truth method exclusively used $1.5$ diameter BBs.

For the first method used to obtain ground truth poses, three CT scans were obtained for each specimen: preoperative without BBs, pre-osteotomy with BBs inserted, and postoperative with BBs inserted, osteotomies performed, and the fragment relocated.
The reference anatomical coordinate frame (APP) was established in the preoperative CT without BBs.
Using a point-to-surface iterative closest point (ICP) registration \cite{besl1992amethod_pami}, the transformation between the preoperative CT without BBs and the pre-osteotomy CT with BBs is estimated.
This uses a pelvis segmentation of the preoperative CT without BBs and points manually digitized on a volume rendering of the pre-osteotomy CT with BBs.
BB locations prior to any osteotomies and fragment relocation are obtained by manually annotating the BBs in the pre-osteotomy CT with BBs and mapping them into the APP.
BBs were also manually annotated in the postoperative CT and correspondences were established.
The transformation between the APP and postoperative CT frame was computed using a paired-point cloud registration \cite{horn1987closed} between the eight ilium BBs in the APP frame and the postoperative CT.
After mapping the postoperative fragment BBs into the APP frame, the relocated pose of the fragment is computed using a paired-point cloud registration between the sets of fragment BBs.

The second ground truth method was devised when a pre-osteotomy CT scan with BBs was not available.
Just as in the first method, the BBs are manually digitized in the postoperative CT.
The postoperative BBs are mapped into the APP frame using a transformation from postoperative CT to preoperative CT computed via an ICP strategy.
3D positions of the fragment BBs, prior to the fragment's relocation, are recovered using three 2D fluoroscopy images collected prior to the osteotomies.
For each of these pre-osteotomy 2D images, all BBs and any available anatomical landmarks are manually digitized.
The mapping from the APP to each projective frame is solved in a modified strategy of the multi-landmark PnP strategy described in section \ref{sec:supp_gen_regi}.
First, the combined set of anatomical landmarks and ilium BBs are used with the POSIT algorithm.
Second, the POSIT pose is used as an initialization to an iterative minimization of re-projection differences using \textit{only} the ilium BBs.
The relative view geometry is recovered using the pelvis coordinate frame, and the pose of the fragment BB constellation is estimated with a multi-view iterative minimization of fragment BB re-projection differences.
Applying this transformation to the fragment BBs yields the pre-osteotomy BB locations.
The same process is repeated for three post-osteotomy fluoroscopic images, and a paired-point cloud registration is used to recover the relative pose of the fragment.
It is important to note that this method uses a mix of 2D and 3D information to find the relative pose of the acetabular fragment, but never relies on an intensity-based registration.
%
%
\section{Results}
\label{sec:supp_results}

\subsection{Simulation Study}
%
\begin{figure}[t]
\centerline{\includegraphics[width=\columnwidth]{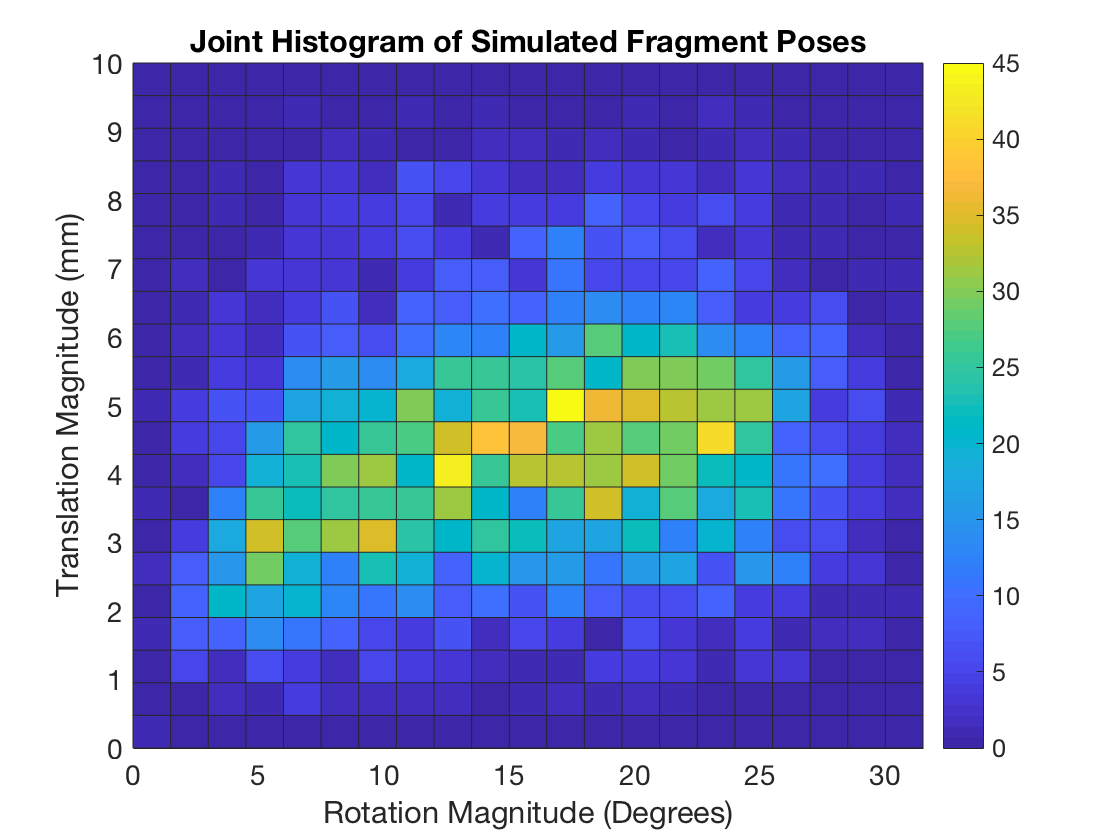}}
\caption{A 2D histogram of the rotation and translation magnitudes for the fragment poses sampled as part of the simulation study.
Note that this is not a uniform distribution matching the parameters in Table \ref{tab:supp_frag_sampling} due to the rejection sampling via collision detection.}
\label{fig:sim_frag_init_hist2d}
\end{figure}
A joint histogram of the simulated fragment pose adjustments is shown in Fig. \ref{fig:sim_frag_init_hist2d}.

\begin{figure}[t]
\centerline{\includegraphics[width=\columnwidth]{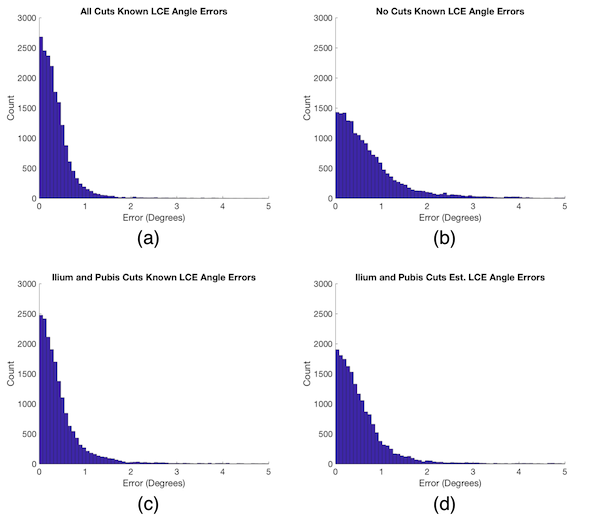}}
\caption{Histograms of lateral center edge (LCE) angle errors for the simulation studies.
(a) The actual fragment shape is known and matches the planned fragment shape,
(b) the actual fragment shape is not known and none of the osteotomies match the planned fragment,
(c) the actual fragment is partially known, with the ilium and pubis osteotomies matching the planned fragment,
(d) the actual fragment is not known, but the ilium and pubis osteotomies are estimated from 2D cut lines.
}
\label{fig:supp_sim_lce_hists}
\end{figure}
With respect to the APP (origin at ipsilateral femoral head), the simulated single-landmark registration initializations had an average offset from ground truth of $5.15 \degree \pm 2.84 \degree$ and $22.50 \text{ mm } \pm 5.74 \text{ mm}$.
After registration, the average offsets from ground truth were
$0.39 \degree \pm 1.29 \degree$ and $2.47 \text{ mm } \pm 5.53 \text{ mm}$,
$0.39 \degree \pm 1.24 \degree$ and $2.40 \text{ mm } \pm 5.47 \text{ mm}$,
$0.40 \degree \pm 1.30 \degree$ and $2.46 \text{ mm } \pm 5.61 \text{ mm}$, and
$0.41 \degree \pm 1.26 \degree$ and $2.48 \text{ mm } \pm 5.37 \text{ mm}$,
for cases when all cuts were known, no cuts were known, the ilium and pubis cuts were known, and the ilium and pubis cuts were estimated (full pelvis registration used), respectively.
When compared to the initialization offsets, the low registration errors indicate that the pelvis registration is robust against the inherent inaccuracy of the single-landmark initialization strategy.

Histograms of the lateral center edge (LCE) angle estimates are shown in Fig. \ref{fig:supp_sim_lce_hists}.
An increase in variance is noticeable as mismatch between the fragment shape used for registration and the actual fragment increases.
%
\subsection{Cadaver Study} \label{sec:supp_results_cad_study}
\begin{figure*}[t]
\centerline{\includegraphics[width=\textwidth]{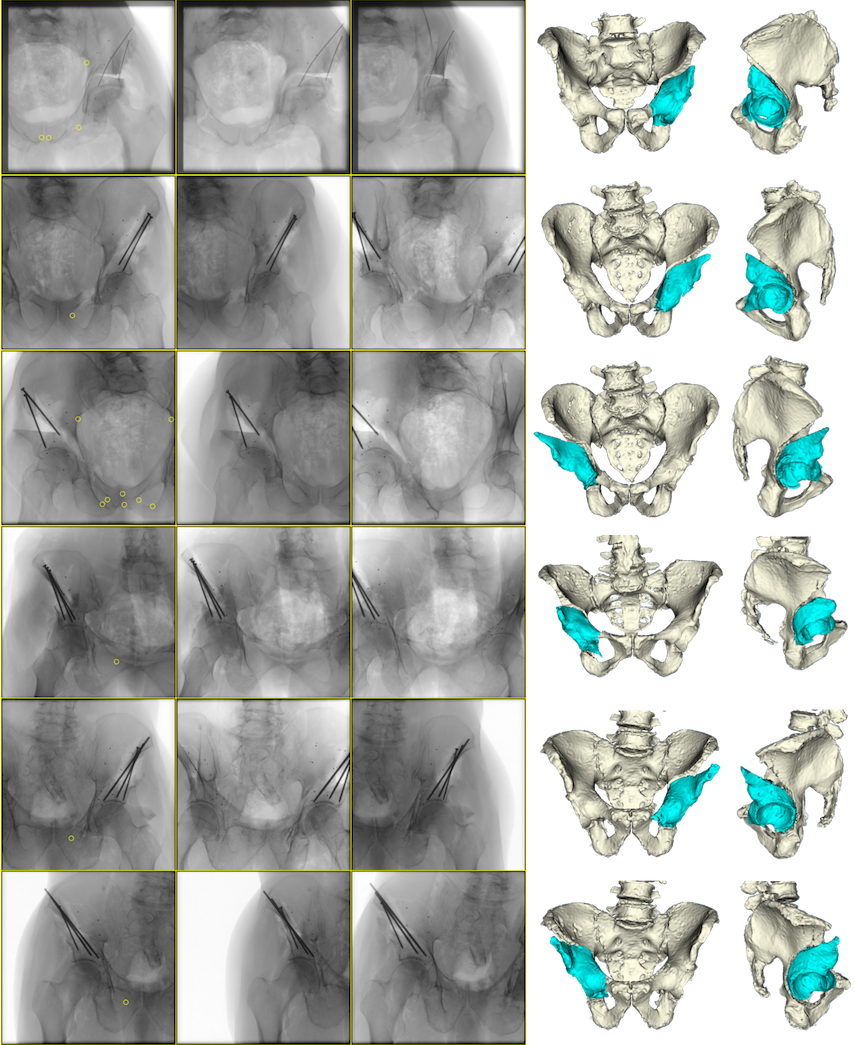}}
\caption{Intraoperative fluoroscopy (left) used in the cadaver studies for pose estimation of the acetabular fragment, along with 3D visualizations of the fragment's pose (right).
Manually annotated landmarks used for registration initialization are overlaid as yellow circles in the approximate AP views shown in the left-most fluoroscopic images.
Note the pelvic tilt away from a standard AP orientation for the two cases that require more than one landmark for initialization.
Registration initialization is automatically obtained for the center and right fluoroscopic views which are pure orbital rotations starting from the AP view.
The second row, from top, is the specimen shown in Figures \ref{fig:frag_view_3d}, \ref{fig:frag_xrays}, \ref{fig:cut_2d_annotate}, and \ref{fig:surgery_view}.
Starting from the top row, the specimen numbers corresponding to Table \ref{tab:supp_preop_ct_params} are 1, 2, 2, 5, 6, and 6.}
\label{fig:all_cad_projs_and_viz}
\end{figure*}
%
\begin{table}[t]
\centering
\caption{Cadaver Surgery Ground Truth Values}
\setlength{\tabcolsep}{3pt}
\begin{tabular}{| l |c|c|c|c|c|c|c|c|c|}
\hline
\multirow{2}{*}{Surgery}  &  \multicolumn{4}{c|}{Rotation ($\degree$)} &  \multicolumn{4}{c|}{Translation (mm)} & \multirow{2}{*}{LCE}  \\ \cline{2-9}
& AP &   LR & IS & Total & AP & LR & IS & Total & \\
\hline
\#1 Left
&  $-4.8$ &  $-0.1$ & $+1.1$ & $5.0$ &  $-0.5$ & $+2.0$ & $-4.3$ & $4.8$ & $36.7$ \\ 
\#2 Left
&  $+2.4$ &  $+13.8$ & $+12.9$ & $18.8$ &  $+1.8$ & $-0.6$ & $-2.1$ & $2.8$ & $39.3$ \\
\#2 Right
&  $+6.8$ &  $+12.3$ & $-14.0$ & $20.2$ & $-1.0$ & $-4.6$ & $-1.0$ & $4.8$ & $42.6$ \\
\#5 Right
&  $+6.2$ &  $+11.5$ & $-6.0$ & $14.6$ & $-0.3$ & $-1.4$ & $-2.0$ & $2.5$ & $57.1$ \\
\#6 Left
&  $-3.1$ &  $+17.2$ & $+10.3$ & $20.5$ & $-2.0$ & $-0.2$ & $-3.1$ & $3.7$ & $42.8$ \\
\#6 Right
&  $+6.3$ &  $+11.6$ & $-1.5$ & $13.3$ & $-1.0$ & $+3.4$ & $+1.9$ & $4.0$ & $41.2$ \\
\hline
\multicolumn{10}{p{245pt}}{Ground truth pose parameters and lateral center edge (LCE) angles of the adjusted fragment for each cadaver surgery.
										 LCE angles are measured in degrees.
										 The transform is with respect to the APP with origin at the ipsilateral femoral head.
										 The specimen numbers correspond to Table \ref{tab:supp_preop_ct_params}.}
\end{tabular}
\label{tab:cadaver_gt_poses}
\end{table}
%
Fig. \ref{fig:all_cad_projs_and_viz} shows the fluoroscopy used for each cadaver experiment, along with 3D visualizations of each fragment's ground truth pose.
The explicit values of the ground truth poses and LCE angles are listed in Table \ref{tab:cadaver_gt_poses}.

For each case that used single-landmark initialization, manually identifying the MOF in each AP view took approximately $2$ seconds.
In order to analyze the annotation times, multiple landmarks were annotated in all initial views, taking an average time of $14.3 \pm 3.3$ seconds.
The times measured for multiple-landmark annotations are listed in Table \ref{tab:cadaver_land_times}.
%
\begin{table}[t]
\centering
\caption{Cadaver Experiment Multiple-Landmark Annotation Times}
\setlength{\tabcolsep}{3pt}
\begin{tabular}{| l | c | c | c | c | c | c |}
\hline
Surgery  &  \#1 Left    &  \#2 Left      &  \#2 Right  & \#5 Right  &  \#6 Left   & \#6 Right  \\ \hline
Time (seconds)      &   8    & 14       &  16   & 17  &  16   &  15  \\
\hline
\multicolumn{7}{p{245pt}}{Time measured to complete the manual annotation of multiple 2D landmarks for the initial views of each surgery.
										 The specimen numbers correspond to Table \ref{tab:supp_preop_ct_params}.
										 When reporting registration results, the multiple landmark initialization strategy was only used for the left side of specimen 1 and right side of specimen 2.}
\end{tabular}
\label{tab:cadaver_land_times}
\end{table}
%

The pipeline components, excluding manual annotations, for each cadaver surgery were re-run 20 times to collect runtime statistics.
Single-landmark pelvis pose initialization is a very quick, non-iterative, method which always completed in less than $0.001$ seconds.
The multiple-landmark pelvis pose initialization was also very quick, and always completed in less than $0.1$ seconds.
Table \ref{tab:cadaver_other_times} shows the times for the intensity-based registrations, manual cut annotations, fragment shape estimations, and multiple-view fragment registrations.
The orbital view registration measurements include time to compute the $181$ DRRs and similarity scores required for initialization.
%
Summing the mean execution times computed for the individual steps of the cadaver study, using single-landmark initialization, and the case of intraoperative fragment estimation, yields an expected runtime of $78.5$ seconds.
When manual annotation is excluded that time is $25.3$ seconds.
This timing analysis was conducted using a single NVIDIA Tesla P100 (PCIe) GPU and access to seven cores of an Intel Xeon E5-2680 v4 CPU.
%
\begin{table}[t]
\centering
\caption{Cadaver Experiment Times for Intensity-Based Registrations, Cut Line Annotation, and Fragment Shape Estimations}
\setlength{\tabcolsep}{3pt}
\begin{tabular}{| p{30pt} | p{35pt} | p{35pt} | p{37pt} | p{35pt} | p{35pt} |}
\hline
\multirow{2}{30pt}{Surgery}     &   \multicolumn{5}{c|}{Time (seconds)} \\ \cline{2-6}
                 &                       Initial View Regi.  &  Orbital View Regis.      &  Cut Line Annotation  &  Frag. Shape Est.  & 3-View Frag. Regi. \\ \hline
 \#1 Left    &                       $4.5 \pm 0.2$      &      $1.5 \pm 0.1$           &              43                  &      $0.7 \pm 0.1$    &  $15.6 \pm 0.4$     \\
 \#2 Left    &                       $4.4 \pm 0.1$      &      $1.6 \pm 0.1$           &              47                  &      $0.9 \pm 0.3$    &  $16.3 \pm 0.6$     \\
 \#2 Right  &                       $3.7 \pm 0.2$      &      $1.6 \pm 0.1$           &              38                 &      $0.9 \pm 0.0$    &   $17.2 \pm 0.6$     \\
 \#5 Right  &                       $5.3 \pm 0.3$      &      $2.2 \pm 0.1$           &              70                 &      $1.2 \pm 0.1$    &   $15.8 \pm 0.5$     \\
 \#6 Left    &                       $4.6 \pm 0.1$      &      $1.6 \pm 0.1$           &               39                &      $1.0 \pm 0.0$    &   $16.6 \pm 0.5$     \\
 \#6 Right  &                       $4.5 \pm 0.3$      &      $1.7 \pm 0.1$           &               70                &      $1.0 \pm 0.2$    &   $16.7 \pm 0.6$     \\ \hline
 All             &                       $4.5 \pm 0.5$      &      $1.7 \pm 0.3$           &    $51.2 \pm 14.9$     &      $1.0 \pm 0.2$    &  $16.4 \pm 0.8$      \\
\hline
\multicolumn{6}{p{245pt}}{Means and standard deviations of runtimes for individual components of the intensity-based registration pipeline for each cadaver surgery.
										Manual cut line annotation was only performed once per surgery.
										For the fragment estimation cases with standard deviation of zero, the actual statistic was $< 0.1$.
										The specimen numbers correspond to Table \ref{tab:supp_preop_ct_params}.}
\end{tabular}
\label{tab:cadaver_other_times}
\end{table}
%

\begin{figure}[t]
\centerline{\includegraphics[width=\columnwidth]{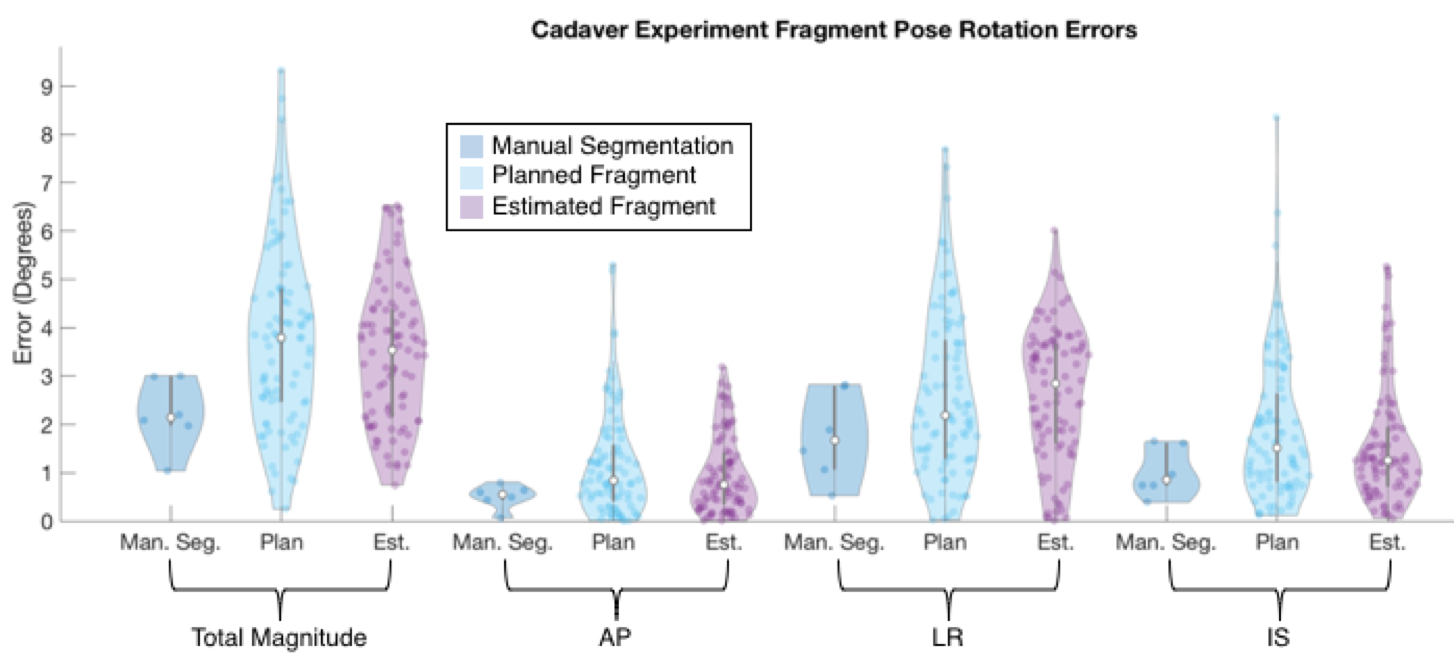}}
\caption{Violin plot of the rotation errors associated with the cadaver experiments.
			 Analogous to Table \ref{tab:cadaver_pose_errors}, the errors are organized by the type of fragment shape used for pose estimation, the total rotation error magnitudes, and the decompositions of the errors about each anatomical axis.
			 For each entry, a scatter plot of the errors is shown by the shaded dots, enclosed by a kernel density function. The white dot indicates the median, and the vertical bar indicates the $[25,75]$ percentiles.}
\label{fig:supp_cad_rot_violin}
\end{figure}

\begin{figure}[t]
\centerline{\includegraphics[width=\columnwidth]{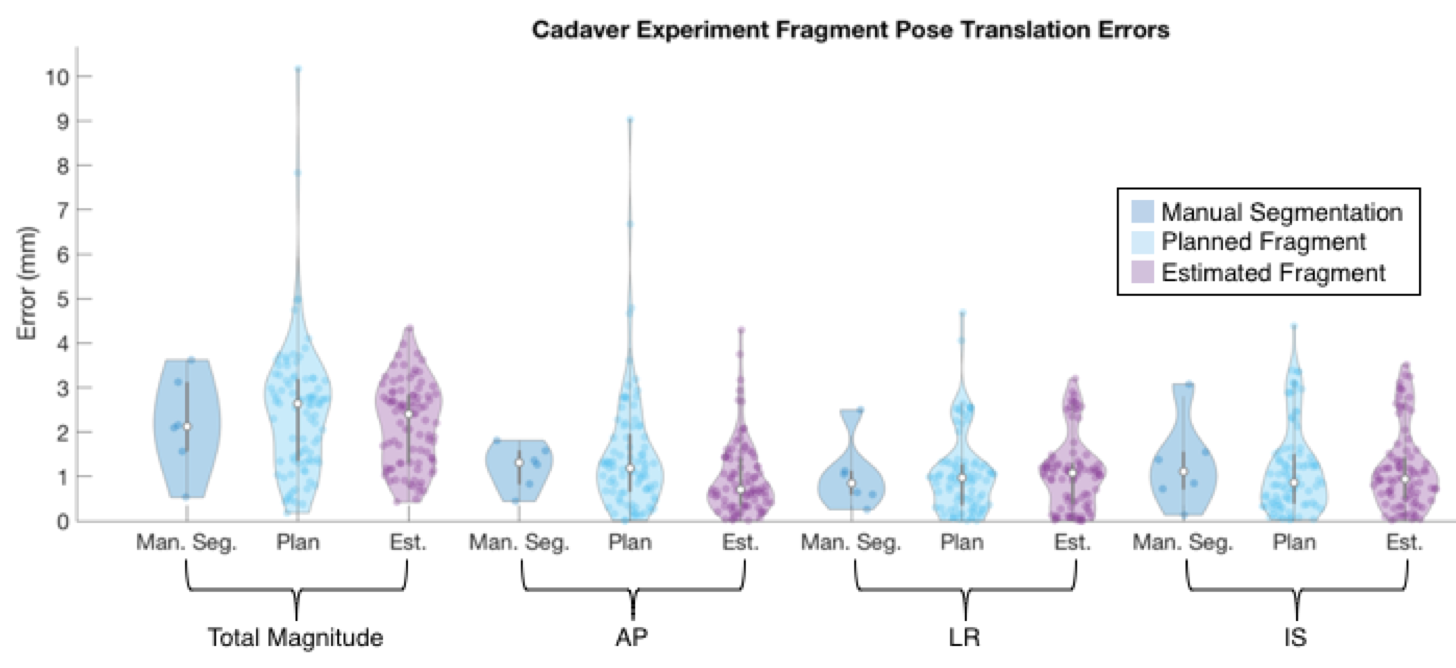}}
\caption{Violin plot of the translation errors associated with the cadaver experiments, organized in the same manner as Fig. \ref{fig:supp_cad_rot_violin}.}
\label{fig:supp_cad_trans_violin}
\end{figure}

\begin{figure}[t]
\centerline{\includegraphics[width=\columnwidth]{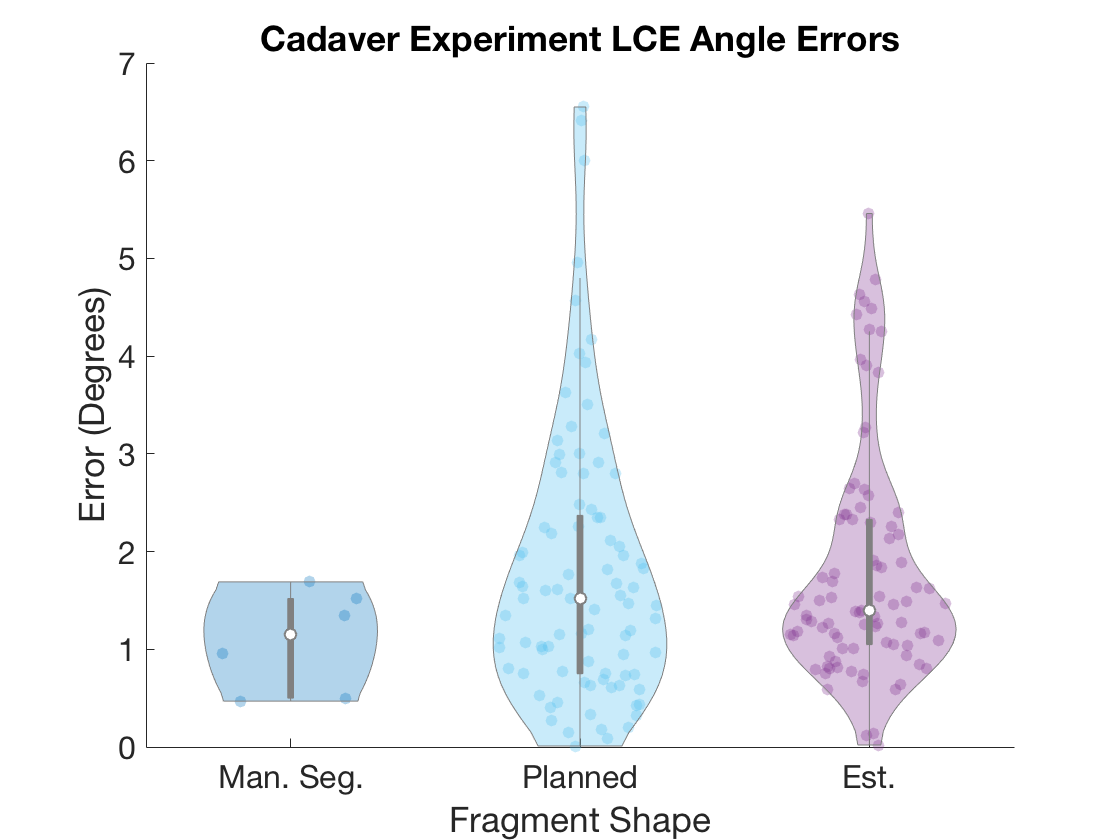}}
\caption{Violin plot of the lateral center edge (LCE) angle errors associated with the cadaver experiments.
		     The errors are organized by the type of fragment shape used during registration.}
\label{fig:supp_cad_lce_violin}
\end{figure}

A breakdown of errors across each surgery, each fragment, each pose dimension, and LCE angles is shown in Table \ref{tab:supp_cadaver_pose_errors}.
Visualizations of the rotation, translation, and LCE angle distributions for all surgeries are shown by the violin plots in Figures \ref{fig:supp_cad_rot_violin}, \ref{fig:supp_cad_trans_violin}, and \ref{fig:supp_cad_lce_violin}, respectively.
%
\begin{table*}[t]
\centering
\caption{Individual Cadaver Surgery Fragment Pose and Lateral Center Edge Angle Errors}
\setlength{\tabcolsep}{3pt}
\begin{tabular}{| p{30pt} | p{42pt} | p{39pt} | p{39pt} | p{39pt} | p{39pt} | p{39pt} | p{39pt} | p{39pt} | p{39pt} | p{39pt} |}
\hline
\multirow{2}{30pt}{Surgery} & \multirow{2}{42pt}{Frag. Shape} &  \multicolumn{4}{c|}{Rotation ($\degree$)} &  \multicolumn{4}{c|}{Translation (mm)}  & \multirow{2}{30pt}{LCE ($\degree$)}  \\ \cline{3-10}
& & AP &   LR & IS & Total & AP &   LR & IS & Total &  \\ 
\hline
\multirow{3}{30pt}{\#1 Left}
 & Manual Seg.        & $0.59$            & $2.79$           & $0.97$           & $3.01$           & $1.58$           & $1.05$           & $3.08$           & $3.62$           & $1.35$            \\
 & Planned          & $0.56 \pm 0.61$  & $3.52 \pm 1.59$ & $2.17 \pm 2.14$ & $4.73 \pm 1.47$ & $2.01 \pm 2.47$ & $1.45 \pm 1.20$ & $2.73 \pm 1.09$ & $4.21 \pm 2.07$ & $1.55 \pm 0.79$  \\
 & Estimated        & $0.29 \pm 0.16$  & $3.71 \pm 0.32$ & $1.14 \pm 0.22$ & $3.90 \pm 0.33$ & $0.75 \pm 0.43$ & $1.12 \pm 0.11$ & $2.86 \pm 0.33$ & $3.19 \pm 0.38$ & $1.30 \pm 0.16$  \\ \hline
\multirow{3}{30pt}{\#2 Left}
 & Manual Seg.        & $0.64$            & $2.82$           & $0.74$           & $2.99$           & $1.27$           & $2.50$           & $1.37$           & $3.12$           & $0.50$            \\
 & Planned          & $1.66 \pm 1.30$  & $2.40 \pm 2.14$ & $1.65 \pm 0.84$ & $3.55 \pm 2.30$ & $1.01 \pm 0.49$ & $2.43 \pm 0.18$ & $0.63 \pm 0.45$ & $2.77 \pm 0.19$ & $1.93 \pm 1.40$  \\
 & Estimated        & $1.39 \pm 0.44$  & $1.84 \pm 1.13$ & $1.50 \pm 0.71$ & $2.93 \pm 0.90$ & $0.64 \pm 0.45$ & $2.64 \pm 0.23$ & $0.36 \pm 0.28$ & $2.79 \pm 0.26$ & $1.80 \pm 0.59$  \\ \hline
\multirow{3}{30pt}{\#2 Right}
 & Manual Seg.        & $0.50$            & $0.53$           & $0.75$           & $1.04$           & $1.82$           & $0.60$           & $0.85$           & $2.09$           & $0.47$            \\
 & Planned          & $1.37 \pm 1.16$  & $2.25 \pm 1.79$ & $2.13 \pm 1.81$ & $3.70 \pm 2.29$ & $1.71 \pm 1.54$ & $0.27 \pm 0.18$ & $0.42 \pm 0.36$ & $1.84 \pm 1.52$ & $1.17 \pm 0.88$  \\
 & Estimated        & $1.25 \pm 0.61$  & $1.32 \pm 1.47$ & $1.09 \pm 0.78$ & $2.30 \pm 1.48$ & $1.26 \pm 1.25$ & $0.28 \pm 0.25$ & $0.35 \pm 0.27$ & $1.44 \pm 1.16$ & $1.14 \pm 0.55$  \\ \hline
\multirow{3}{30pt}{\#5 Right}
 & Manual Seg.        & $0.07$            & $1.45$           & $1.65$           & $2.20$           & $0.44$           & $0.26$           & $0.14$           & $0.53$           & $1.52$            \\
 & Planned          & $0.46 \pm 0.43$  & $2.55 \pm 1.74$ & $1.19 \pm 0.84$ & $2.96 \pm 1.80$ & $0.58 \pm 0.46$ & $0.24 \pm 0.19$ & $0.64 \pm 0.45$ & $0.99 \pm 0.51$ & $1.56 \pm 1.26$  \\
 & Estimated        & $0.31 \pm 0.23$  & $2.47 \pm 1.47$ & $1.12 \pm 0.58$ & $2.80 \pm 1.46$ & $0.61 \pm 0.42$ & $0.17 \pm 0.13$ & $0.93 \pm 0.19$ & $1.18 \pm 0.31$ & $1.70 \pm 1.12$  \\ \hline
\multirow{3}{30pt}{\#6 Left}
 & Manual Seg.        & $0.44$            & $1.07$           & $1.61$           & $1.98$           & $0.83$           & $1.13$           & $0.72$           & $1.57$           & $1.69$            \\
 & Planned          & $2.30 \pm 1.11$  & $2.42 \pm 1.29$ & $3.20 \pm 0.79$ & $4.84 \pm 0.98$ & $1.92 \pm 0.82$ & $1.20 \pm 0.13$ & $1.23 \pm 0.36$ & $2.66 \pm 0.59$ & $3.70 \pm 1.35$  \\
 & Estimated        & $2.15 \pm 0.64$  & $3.28 \pm 0.83$ & $3.74 \pm 0.98$ & $5.51 \pm 0.68$ & $1.82 \pm 0.85$ & $1.19 \pm 0.17$ & $1.14 \pm 0.26$ & $2.54 \pm 0.62$ & $3.87 \pm 0.89$  \\ \hline
\multirow{3}{30pt}{\#6 Right}
 & Manual Seg.        & $0.80$            & $1.88$           & $0.39$           & $2.08$           & $1.36$           & $0.64$           & $1.55$           & $2.16$           & $0.96$            \\
 & Planned          & $0.71 \pm 0.28$  & $2.36 \pm 1.55$ & $1.00 \pm 0.65$ & $2.83 \pm 1.37$ & $1.67 \pm 1.01$ & $0.87 \pm 0.23$ & $1.31 \pm 0.66$ & $2.40 \pm 0.98$ & $0.86 \pm 0.40$  \\
 & Estimated        & $0.50 \pm 0.27$  & $3.03 \pm 1.17$ & $0.72 \pm 0.53$ & $3.26 \pm 0.98$ & $0.77 \pm 0.55$ & $1.14 \pm 0.33$ & $1.21 \pm 0.51$ & $1.95 \pm 0.45$ & $1.02 \pm 0.37$  \\ \hline
\multicolumn{11}{p{490pt}}{Summary of the means and standard deviations of the fragment pose and lateral center edge angle (LCE) errors for each cadaver surgery.
										 Specimen numbers correspond to those in Table \ref{tab:supp_preop_ct_params}.
										 The organization presented here is analogous to Table \ref{tab:cadaver_pose_errors}.}
\end{tabular}
\label{tab:supp_cadaver_pose_errors}
\end{table*}
%
%
%
\end{document}